\documentclass[]{icas2026}

\usepackage[english]{babel}
\usepackage{amsmath}
\usepackage{amssymb}
\usepackage{amsfonts}
\usepackage{algorithmic}
\usepackage{graphicx}
\usepackage{textcomp}
\usepackage[dvipsnames]{xcolor}
\definecolor{DLRBlack}{gray}{0}
\definecolor{DLRGrey}{gray}{0.420} 
\colorlet{DLRGray}{DLRGrey}
\definecolor{DLRWhite}{gray}{1}

\colorlet{DLREagleColor}{DLRBlack}
\colorlet{DLRTextColor}{DLRBlack}

\colorlet{DLRDarkerGrey}{DLRGrey}
\definecolor{DLRDarkGrey}{gray}{0.537} 
\definecolor{DLRMediumGrey}{gray}{0.702} 
\definecolor{DLRLightGrey}{gray}{0.820} 
\definecolor{DLRLighterGrey}{gray}{0.929} 

\colorlet{DLRDarkerGray}{DLRDarkerGrey}
\colorlet{DLRDarkGray}{DLRDarkGrey}
\colorlet{DLRMediumGray}{DLRMediumGrey}
\colorlet{DLRLightGray}{DLRLightGrey}
\colorlet{DLRLighterGray}{DLRLighterGrey}

\definecolor{DLRDarkerBlue}{RGB}{0, 106, 144}
\definecolor{DLRDarkBlue}{RGB}{0, 156, 208}
\definecolor{DLRBlue}{RGB}{33, 187, 223}
\colorlet{DLRMediumBlue}{DLRBlue}
\definecolor{DLRLightBlue}{RGB}{149, 212, 238}
\definecolor{DLRLighterBlue}{RGB}{201, 232, 251}

\definecolor{DLRDarkerGreen}{RGB}{115, 163, 63}
\definecolor{DLRDarkGreen}{RGB}{158, 193, 76}
\definecolor{DLRGreen}{RGB}{199, 214, 84}
\colorlet{DLRMediumGreen}{DLRGreen}
\definecolor{DLRLightGreen}{RGB}{215, 223, 116}
\definecolor{DLRLighterGreen}{RGB}{228, 234, 173}

\definecolor{DLRDarkerYellow}{RGB}{224, 177, 57}
\definecolor{DLRDarkYellow}{RGB}{254, 206, 73}
\definecolor{DLRYellow}{RGB}{255, 223, 73}
\colorlet{DLRMediumYellow}{DLRYellow}
\definecolor{DLRLightYellow}{RGB}{255, 234, 117}
\definecolor{DLRLighterYellow}{RGB}{255, 248, 189}

\definecolor{DLRDarkestBlue}{RGB}{0, 50, 69}
\definecolor{DLRDarkestGreen}{RGB}{99, 119, 34}
\definecolor{DLRDarkestYellow}{RGB}{190, 150, 0}

\definecolor{DLRRed}{RGB}{179, 63, 61}
\colorlet{DLRMediumRed}{DLRRed}
\definecolor{DLRLightRed}{RGB}{199, 122, 109}
\definecolor{DLRLighterRed}{RGB}{223, 180, 168}

\definecolor{DLRDarkestGray}{gray}{0.302} 
\usepackage{soul}
\usepackage{xfrac}

\usepackage{csquotes}

\usepackage{microtype}

\usepackage{tabularx}
\usepackage{booktabs}
\usepackage{tabularray}
\usepackage{longtable}
\usepackage{array}
\usepackage{ragged2e}
\usepackage{makecell}
\newcolumntype{L}[1]{>{\RaggedRight\arraybackslash}p{#1}}
\UseTblrLibrary{booktabs}

\usepackage{placeins}

\usepackage{hyperref}
\usepackage[capitalize, noabbrev]{cleveref}

\usepackage{listings}

\colorlet{jsonBg}{DLRWhite}
\colorlet{jsonString}{DLRDarkestGray}
\colorlet{jsonKeyword}{DLRRed} 
\colorlet{jsonNumber}{DLRDarkBlue} 
\colorlet{jsonPunct}{DLRDarkGreen} 
\colorlet{jsonGray}{DLRMediumGray} 

\lstdefinelanguage{json}{
  basicstyle = \small\ttfamily\color{jsonNumber},
  numbers = left,
  numberstyle = \scriptsize\color{jsonGray},
  stepnumber = 1,
  numbersep = 10pt,
  string = [s]{"}{"},
  stringstyle = \color{jsonString},
  morekeywords = {true, false, null},
  keywordstyle = \color{jsonKeyword}\bfseries,
  showstringspaces = false,
  breaklines = true,
  frame = false,
  framerule = 0.4pt,
  rulecolor = \color{lightgray},
  backgroundcolor = \color{jsonBg},
  literate =
    *{:}  {{{\color{jsonPunct} {:}}}}{1}
     {,}  {{{\color{jsonPunct} {,}}}}{1}
     {\{} {{{\color{jsonPunct} {\{}}}}{1}
     {\}} {{{\color{jsonPunct} {\}}}}}{1}
     {[}  {{{\color{jsonPunct} {[}}}}{1}
     {]}  {{{\color{jsonPunct} {]}}}}{1},
}

\usepackage{siunitx}
\DeclareSIUnit\knot{kn}
\DeclareSIUnit\nauticalmile{NM}
\DeclareSIUnit\foot{ft}
\DeclareSIUnit\flops{FLOPS}

\usepackage{bbding}
\usepackage{tikz}
\usepackage[edges]{forest}
\usepackage{wheelchart}
\usetikzlibrary{
    calc,
    backgrounds,
    bbox,
    decorations.text,
    shapes.geometric,
    arrows,
    arrows.meta,
    angles,
    math,
    patterns,
    quotes
}

\usepackage{pgfplots}
\usepackage{pgfplotstable}
\usepgfplotslibrary{groupplots}
\pgfplotsset{compat=newest}
\usepgfplotslibrary{
    fillbetween,
    statistics
}

\makeatletter

\def\@seccntformat#1{%
  \csname the#1\endcsname
  \csname ICAS@secdot@#1\endcsname
  \quad
}
\expandafter\def\csname ICAS@secdot@section\endcsname{.}
\makeatother

\TitlePaper{On the Applicability of Safety Nets: A Safety-By-Design Solution for Certifying Neural Networks}
\AuthorPaper[1]{Johann Maximilian Christensen}
\AuthorPaper[2]{Thomas Stefani}
\AuthorPaper[1]{Elena Hoemann}
\AuthorPaper[1]{Frank Köster}
\AuthorPaper[1]{Sven Hallerbach}
\affil[1]{%
\textit{Institute for AI Safety and Security}\\
\textit{German Aerospace Center (DLR)}\\
Sankt Augustin, Germany%
}
\affil[2]{%
\textit{Institute for AI Safety and Security}\\
\textit{German Aerospace Center (DLR)}\\
Ulm, Germany%
}

\abstractEnglish{%
The integration of Artificial Intelligence (AI) in safety-critical aviation systems presents significant challenges for certification and deployment.
Aviation, often regarded as the safest form of transportation, relies on numerous safety-critical systems, from flight control to collision avoidance.
For future safety-critical AI-based systems, EASA requires a Safety-by-Design approach, which can be achieved by using Safety Nets that combine neural network compression with lookup tables to ensure \qty{100}{\percent} correct runtime behavior across the discretized operational design domain.
Although Safety Nets have been studied, no comprehensive study of their performance characteristics and system design trade-offs has been conducted, leaving critical questions unanswered.
This work presents the first systematic analysis of the trade-off between neural network and lookup table size in Safety Nets for next-generation collision-avoidance systems.
By systematically comparing neural networks with diverse architectures---including varying activation functions (ReLU, LeakyReLU, GELU), hidden-layer configurations, and encoding strategies---this study identifies optimal design parameters that minimize overall storage and memory requirements while maintaining certification compliance.
Results demonstrate that architectures with 3 to 5 hidden layers, each with approximately 50 to 100 nodes, combined with one-hot encoding, achieve the best balance.
In these configurations, neural networks accurately represent at least \qty{97}{\percent} of the data, while compact lookup tables handle the remaining errors.
The resulting Safety Nets reduce the system size by almost three orders of magnitude for HCAS and more than one order of magnitude for VCAS, fitting within the memory budget of current avionics hardware while guaranteeing \qty{100}{\percent} correct outputs across the entire discretized input space, as required by EASA guidelines.
This work provides the first-ever open-source implementation of Safety Nets for HCAS and VCAS with replicable results, demonstrating a practical pathway toward certifiable AI-based systems in aviation and establishing Safety Nets as a viable Safety-by-Design solution for safety-critical applications.
}
\keywords{AI Engineering, Safety-by-Design, Artificial Intelligence, Neural Networks, AI Certification}

\begin{document}

\body  

\section{Introduction}
Artificial Intelligence (AI) has seen increasing adoption across many domains, including aviation.
The estimated annual growth rate for AI-based applications in aviation of \qty{35}{\percent} together with the shortage of approximately \num{700000} pilots projected through 2043, will inevitably lead to increased use of safety-critical AI-based automation in the cockpit to reduce pilot workload~\cite{PrecedenceResearch2022, Boeing2024}.
One system expected to reduce pilots' workload is the future Airborne Collision Avoidance System X (ACAS~X), intended to be a drop-in replacement for its predecessor, the Traffic Collision Avoidance System II (TCAS II), which often generates false-positive alerts, unnecessarily increasing workload~\cite{ED-143, ED-256, ED-275, DO-385, DO-386}.
Implementing ACAS~X, however, faces a significant obstacle as the newly designed system cannot run on current avionics hardware.
The raw dynamic-programming output of the underlying Markov decision processes (MDP) requires hundreds of gigabytes of memory; even the horizontal ACAS~Xu logic table, downsampled by a factor of approximately 180, still exceeds \qty{2}{\giga\byte}, and at least \qty{4}{\gibi\byte} of memory would be required to embed the system~\cite{Julian2019a, Damour2021}.
Current avionics hardware, in contrast, offers memory on the order of hundreds of megabytes---the latest variants of the GE Aerospace Flight Management Computer provide \qty{512}{\mega\byte} of RAM~\cite{GEC2018}---roughly one order of magnitude less than the already heavily reduced table and roughly three orders of magnitude less than the raw MDP output.
The full-specification state spaces of ACAS~Xa and ACAS~Xu, defined over the complete operational envelopes of ED-256/DO-385 and ED-275/DO-386~\cite{ED-256, ED-275, DO-385, DO-386}, are larger still by up to four orders of magnitude than the proof-of-concept HCAS and VCAS tables compressed in this work (cf.\ \cref{sec:Discussion} and \cref{tab:spec-coverage}).
Thus, research has focused on compressing data using neural networks, leading to an AI-based system for a safety-critical aviation application~\cite{Christensen2024}.
Whenever AI-based systems are used in safety-critical applications, a Safety-by-Design development approach must be employed to ensure successful subsequent certification~\cite{Hoemann2026}.
Deploying any AI-based system in a safety-critical application in aviation requires the development process to be compliant with EASA's guidelines, which are intended to mitigate erroneous outputs~\cite{EUASA2024, Christensen2025, Werner2026}.
For ACAS~X, an incorrect advisory issued to the pilots could lead to catastrophic emergencies.
Thus, a correct learned representation of the MDP is paramount.
Moreover, for ACAS~X, specifically the simplified open-source HCAS and VCAS~\cite{Julian2019}, there exist different approaches to ensure a correct representation.
These range from closed-loop verification properties~\cite{ManzanasLopez2023} to reachability analyses~\cite{Julian2019, Katz2017} to tools that inherently guarantee safety, such as Safety Nets~\cite{Damour2021, Christensen2024}.
Only the latter fully adheres to a Safety-by-Design approach, as it incorporates built-in protection against erroneous outputs, exceeding EASA's requirements.
Safety Nets combine the compression capabilities of neural networks with the flawless representation of lookup tables: neural networks store the bulk of the data, while the lookup table stores only input vectors for which the neural network is known to produce incorrect outputs.
Because Safety Nets provide an additional layer of protection against errors arising from misrepresented data via lookup tables, a better-performing neural network requires smaller lookup tables.
Conversely, even the poorest performing neural network can be made safe at the cost of a larger lookup table.
Thus, a balance must be achieved to minimize the overall system size in terms of storage and memory requirements while ensuring acceptable execution times, thereby enabling deployment of the combined system on current avionics hardware.
This paper explores the applicability of Safety Nets as a Safety-by-Design solution for certifiable neural networks while ensuring compatibility with the W-shaped development process required by EASA.
While prior works~\cite{Damour2021, Christensen2024} already conducted preliminary studies on the performance of Safety Nets for HCAS and VCAS, this work provides a more systematic analysis of the actual performance of Safety Nets.
To achieve this, various neural network architectures are investigated, in contrast to prior work, which largely reused the general neural network architecture proposed in~\cite{Julian2016}.
Recent work reports that LeakyReLU often outperforms ReLU on general classification and regression benchmarks~\cite{Dubey2019, Maas2013, Xu2015}.
Finally, even minor deviations in the number and size of the hidden layers can drastically improve the performance of neural networks.
However, while a better-performing neural network is an improvement nonetheless, this is not strictly required to follow a Safety-by-Design approach when utilizing Safety Nets.

The paper is structured as follows: \cref{sec:soa} gives an overview of the state-of-the-art in terms of AI-based applications in aviation and how to develop them in a safety-by-design manner.  
Next, the overarching use case for the paper is presented in \cref{sec:UseCase}, followed by a high-level overview of how an AI-based system for this use case would be developed in accordance with EASA's concept paper~\cite{EUASA2024}.
Afterward, focusing on the development of Safety Nets, a performance study is presented, investigating trade-offs between different hyperparameters, followed by the presentation of the final Safety Net in \cref{sec:SafetyNet}.
Next, in \cref{sec:Certification}, the concept of Safety Nets is matched with objectives from the EASA concept paper, highlighting how Safety Nets can streamline the certification process.
Finally, in \cref{sec:Discussion}, the results of the paper are discussed and, in \cref{sec:Conclusion}, conclusions are drawn.

\section{State of the Art}\label{sec:soa}
The development of AI-based systems for safety-critical applications sits at the intersection of three active research areas: advances in neural network design, the application of AI to aviation, and the certification of AI-based systems in regulated environments.
Each is reviewed in turn, with a particular focus on the ACAS~X use case and the Safety Net methodology that forms the core of this work.

\subsection{Artificial Intelligence in Aviation}
The potential of AI-based systems in aviation has been recognized across a broad range of operational domains, ranging from operational decision support to perceptual tasks.
Perceptual applications include autonomous visual landing guidance~\cite{EASADaedalean2020, EASADaedalean2024}, runway detection from image data~\cite{Ducoffe2023, Mussot2026}, and foreign object debris detection on runway surfaces~\cite{Papadopoulos2021, Taupik2023}, while decision-support applications include air-traffic management~\cite{Stefani2024, Berro2025}.

A recurring challenge across all of these applications is the mismatch between the iterative, data-driven development process typical of machine learning and the document-centric, phase-gated development assurance processes prescribed by aviation standards such as DO-178C and ARP4754A~\cite{DO-178C, SAE-ARP4754A}.
A broad survey of AI systems in aviation identifies this regulatory incompatibility as a central obstacle to deployment~\cite{Kashyap2019}, while a systematic literature review of certification approaches for safety-critical Machine Learning (ML) systems concludes that no single method is sufficient across all criticality levels, with formal verification, runtime monitoring, and architectural constraints each addressing only a subset of certification objectives~\cite{Tambon2022}.
A more agile, DevOps-oriented development paradigm has therefore been proposed as a complement to the V-model, accommodating the iterative nature of ML development while remaining compatible with certification requirements~\cite{Christensen2025, Werner2026, App2024, Stefani2024a, Stefani2024b}.

Compatibility of ML-based systems with conventional development assurance processes has been demonstrated up to DAL~C for select applications, including runway sign classification~\cite{Dmitriev2023} and low-criticality airborne systems more generally~\cite{Dmitriev2021, Sridhar2025}.
At higher criticality levels, however, standard verification and validation methods are insufficient, and dedicated AI-specific assurance frameworks are required~\cite{Zaeske2023}.
Dataset management and Operational Design Domain (ODD) compliance have also been identified as prerequisites for assurance, with recent work addressing how datasets can be constructed to remain compliant with a system's ODD~\cite{Cappi2024} and how unintended model behaviors can be detected prior to deployment~\cite{Spaeth2024}.

\subsection{Certification of AI-Based Systems}
In response to the challenges outlined above, both EASA and the FAA have issued initial guidance documents detailing their requirements for the future certification of AI-based systems in aviation~\cite{EUASA2024, FAA2024}.
EASA's concept paper introduces the W-shaped development process---an augmentation of the classical V-model with dedicated phases for AI/ML model training, learning assurance, and model verification---and defines a set of learning assurance and implementation objectives that applicants must satisfy~\cite{EUASA2024}.
These objectives place particular emphasis on the representativeness of the trained model, requiring quantifiable generalization bounds, requirements-based verification against a formal specification, and evidence that model behavior is stable and robust across the entire ODD, including edge cases and corner cases~\cite{EUASA2024}.

Satisfying these objectives through statistical testing alone is insufficient, as no finite test set can certify correctness across a complete, potentially large input space~\cite{EUASA2024}.
This problem has motivated research into formal verification methods for neural networks.
Reluplex~\cite{Katz2017} was the first satisfiability-modulo-theories (SMT)-based solver specifically designed for verifying deep neural networks, and was applied directly to early ACAS~Xu neural networks to prove a set of ten avoidance meta-properties.
Reluplex demonstrated that formal verification of neural networks is tractable for small networks, but its computational cost scales exponentially with network size, limiting its practical applicability to the verification of bounded local regions rather than global policy agreement.
Reachability-based approaches have since been developed as more scalable alternatives: star-set reachability analysis has been applied to verify that neural network compression preserves the advisory structure of ACAS~Xu lookup tables~\cite{ManzanasLopez2021}, and a subsequent comparative evaluation of multiple verification methods for air-to-air collision avoidance concluded that no existing tool offers both completeness and scalability simultaneously, with each method producing a different coverage-runtime trade-off~\cite{ManzanasLopez2023}.
Taken together, these results indicate that formal verification methods can provide strong guarantees for selected meta-properties but have not been applied to the problem of full policy agreement across a complete, discretized input space---the requirement that drives the Safety Net approach.

\subsection{Neural Network Design and Activation Functions}
For the specific problem of compressing discrete, tabular functions derived from Markov decision processes, fully connected feedforward networks have been the architecture of choice~\cite{Julian2019a, Damour2021, Julian2016}.
A critical design choice for such networks is the activation function, which governs the nonlinear transformation applied at each hidden node and thereby shapes the network's representational capacity and training dynamics.

The Rectified Linear Unit (ReLU) has been widely adopted as the default activation function due to its computational simplicity and its ability to mitigate vanishing gradients during training~\cite{Zhang2024a}.
However, ReLU suffers from the \emph{dying ReLU} problem, in which neurons with consistently negative pre-activations produce zero gradients and cease to contribute to learning.
LeakyReLU was introduced to address this by assigning a small, fixed slope to negative inputs, and has been shown empirically to outperform ReLU on a variety of classification and regression benchmarks~\cite{Dubey2019, Maas2013, Xu2015}.
The Gaussian Error Linear Unit (GELU), which weights inputs by their cumulative Gaussian probability~\cite{Hendrycks2016}, has become the de facto activation function in large-scale transformer-based language and vision models~\cite{Devlin2019}, where its smooth nonlinearity improves optimization over complex, high-dimensional loss landscapes.
Despite these general trends, the relative performance of activation functions is task-dependent, and their interaction with the discrete, bounded structure of MDP-derived policy tables has not previously been studied in the context of Safety Nets.
Prior work on neural network compression for ACAS~X evaluated only the ReLU activation function~\cite{Julian2019a, Damour2021, Julian2016}, leaving the question of whether alternative activations yield smaller combined system sizes entirely open.

\subsection{Neural Network Compression for ACAS~X}
Neural network-based compression of ACAS~X was first proposed to replace the multi-gigabyte MDP lookup tables with an ensemble of small fully connected networks~\cite{Julian2016}, and subsequently extended to deeper architectures~\cite{Julian2019a}.
Early reachability-based safety guarantees for these networks proved closed-loop avoidance properties for the open-source HCAS and VCAS implementations that also serve as the use case for the present work~\cite{Julian2019}.
Architectural challenges arising from the integration of such neural network-based systems into real avionics platforms are discussed in current research~\cite{Janson2023}, and related compression approaches have been explored for small uncrewed aircraft collision avoidance~\cite{Irfan2020}.
A broader pathway toward real-world deployment, encompassing System Engineering, DevOps, and hardware integration aspects, has been investigated in subsequent work~\cite{Stefani2024b, Christensen2024a}.

\subsection{Safety Nets}
Safety Nets were introduced as a hybrid architecture that combines the compression efficiency of neural networks with the provable correctness of lookup tables, applied to the horizontal conflict resolution component of ACAS~Xu~\cite{Damour2021}.
The input state space is partitioned into a set of multi-dimensional boxes whose boundaries are defined by the lookup table (LUT) parameter values; for each box, a formal verification tool checks whether the neural network's reachable advisories are consistent with those of the LUT\@.
Boxes for which this property cannot be confirmed are added to the safety net, which stores the corresponding LUT entries.
At inference time, a check module consults the safety net first; if the current input falls within a stored box, the LUT entry is returned, otherwise the neural network is queried.
The combined system is thereby guaranteed to agree with the formal LUT specification across the entire operational domain.
Verification was performed using the abstract interpretation tool DeepPoly~\cite{Singh2019} as a first pass, with the SMT solvers Reluplex and Planet invoked as a fallback on the boxes DeepPoly left undecided; the procedure was restricted to \(\tau = \qty{0}{\second}\) and constant ownship and intruder speeds of \qty{438}{\foot\per\second} and \qty{414}{\foot\per\second}, partitioning the reduced state space into \num{304000} three-dimensional boxes.
Only the ReLU activation function was evaluated, and the study compared regression against classification targets as well as regular against decreasing layer architectures, selecting the two configurations that best balanced accuracy with memory footprint.
Crucially, the implementation details required to reproduce these results were not disclosed.

The certification implications of this hybrid architecture were subsequently formalized~\cite{Gabreau2022}, where the hybrid architecture is situated within the assurance case framework developed by the EUROCAE WG-114/SAE~G-34 joint working group.
That work structures the full certification argumentation for the ACAS~Xu hybrid controller using Goal Structuring Notation, mapping evidence artifacts from the ML development process---including formal verification of generalization and performance measurements---to the learning assurance objectives of the forthcoming AS6983 standard.
The certification strategy rests on the same core argument as the Safety Net construction itself.
Because the LUT constitutes the formal specification, exact agreement between the hybrid controller's outputs and the LUT decisions constitutes exhaustive requirements-based verification, and any residual disagreement is explicitly captured and corrected by the safety net.
However, no implementation artifacts were released, and the performance study was not extended beyond the architectures and activation functions already evaluated.
For context, the prior Safety Net study reported a horizontal-only hybrid system whose footprint was dominated by the neural network rather than the LUT.
The best configurations occupied roughly \qty{122}{\mega\byte} (regular width, \num{50} nodes per layer) to \qty{218}{\mega\byte} (decreasing width) of network parameters, against Safety Net extracts of only \qty{162}{\kilo\byte} and \qty{39}{\kilo\byte}, respectively, with neural-network--LUT agreement rates of \qty{95.4}{\percent} and \qty{95.8}{\percent}~\cite{Damour2021}.
This inverts the balance observed for VCAS in the present work, where the lookup table dominates; the difference stems from their use of large regression networks and a coarser box partition, whereas the present work evaluates a finer, exhaustively swept discrete grid across all \(\tau{}\) values and both systems.
Note, however, that the box-based partition of prior studies~\cite{Damour2021} certifies agreement over continuous regions of the input space, whereas the exhaustive sweep employed in this work certifies agreement point-wise on the discretized grid; the implications of this distinction for runtime operation are discussed in \cref{sec:SafetyNet}.

The present work builds on this foundation by providing the first systematic analysis of how neural network architecture---including activation function, depth, width, and output encoding---influences the combined size of the Safety Net, and by releasing the first publicly reproducible Safety Net implementation for both HCAS and VCAS\@.
Rather than treating the neural network architecture as fixed and focusing exclusively on the certification argumentation, this work treats the architecture itself as a design variable, with the goal of identifying configurations that minimize total system size while satisfying EASA's requirements by construction.

\section{The ACAS X Use Case}\label{sec:UseCase}
ACAS~X, compared to TCAS II, not only generates vertical Resolution Advisories (RAs) with ACAS~Xa but also horizontal RAs with ACAS~Xu.
While ACAS~Xa is meant as a drop-in replacement for TCAS II, ACAS~Xu, with its additional horizontal advisories, is designed for Unmanned Aircraft Systems (UAS)~\cite{ED-256, ED-275, DO-385, DO-386}.
Both, however, use a complex MDP to generate a \(Q\)-value for every possible advisory, given the current states of the ownship and intruder, and then select the resolution advisory with the highest associated value.
The \(Q\)-value thus represents a state-action score, quantifying the desirability of issuing a particular advisory from a given state; the advisory with the highest \(Q\)-value is selected as the optimal action for that state.
In safety-critical situations, an incorrect advisory could command the pilot to maneuver toward rather than away from the intruder, potentially resulting in a near mid-air collision (NMAC)~\cite{ED-256, ED-275, Julian2019}.

\begin{figure}[htb]
    \begin{subfigure}[t]{0.47\linewidth}
        \centering
        \includegraphics{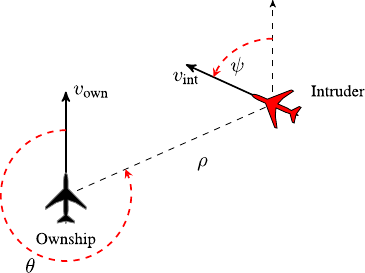}
        \caption{Geometry of the horizontal collision avoidance scenario for HCAS, from~\cite{Julian2019}. The black ownship is trying to avoid the (malicious) red intruder by diverting to the left or right.}
        \label{fig:HCAS}
    \end{subfigure}
    \hfill
    \begin{subfigure}[t]{0.48\linewidth}
        \centering
        \includegraphics{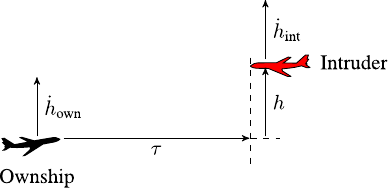}
        \caption{Geometry of the vertical collision avoidance scenario for VCAS, from~\cite{Julian2019}. The black ownship is trying to avoid the (malicious) red intruder by diverting through climbing or descending.}
        \label{fig:VCAS}
    \end{subfigure}
    \caption{Overview of the general geometry of both HCAS (\cref{fig:HCAS}) and VCAS (\cref{fig:VCAS}).}
    \label{fig:CAS}
\end{figure}

To reduce the overall system size, particularly of the threat resolution module, which provides horizontal and vertical RAs, a neural network-based compression, called HCAS and VCAS for horizontal and vertical collision avoidance, respectively, was proposed~\cite{Julian2019}.
Both systems are open-source, proof-of-concept implementations inspired by early prototypes of ACAS~Xa and ACAS~Xu, and serve as the use case for this work~\cite{Julian2019}.
The general encounter geometries for HCAS and VCAS are shown in \cref{fig:HCAS,fig:VCAS}, respectively.

\subsection{Horizontal Collision Avoidance System (HCAS)}
HCAS issues horizontal turning advisories to the ownship to avoid an NMAC with an intruder aircraft.
As shown in \cref{fig:HCAS}, the range describes the encounter to the intruder \(\rho\), the bearing angle \(\theta\) of the intruder relative to the ownship heading, the relative heading angle of the intruder \(\psi\), the ownship speed \(v_\mathrm{own}\), the intruder speed \(v_\mathrm{int}\), the time to the closest point of approach \(\tau\), and the previous advisory \(s_\mathrm{adv}\).
The complete parameter ranges used in this work are listed in \cref{tab:HCAS_ODD} and correspond to the open-source implementation\footnote{Available at \url{https://github.com/sisl/HorizontalCAS}.}~\cite{Julian2019}.
Note that ownship and intruder speeds are held constant, as the original implementation fixes both to \qty{200}{\foot\per\second}.
The five possible advisories are Clear of Conflict (COC), Weak Left (WL), Weak Right (WR), Strong Left (SL), and Strong Right (SR), corresponding to \(s_\mathrm{adv} \in \{0, 1, 2, 3, 4\}\).

Because the two continuous angular variables \(\theta\) and \(\psi\), as well as \(\tau\), would yield an intractably large monolithic input space, HCAS is decomposed into an ensemble of 40 individual neural networks.
Each network corresponds to one combination of the discretized time to closest point of approach \(\tau \in \{\qty{0}{\second}, \qty{5}{\second}, \qty{10}{\second}, \qty{15}{\second}, \qty{20}{\second}, \qty{30}{\second}, \qty{40}{\second}, \qty{60}{\second}\}\) and one of the five possible previous advisories \(s_\mathrm{adv}\), resulting in \(8 \times 5 = 40\) subsystems~\cite{Julian2019, Julian2016}.
Thus, each subsystem takes only three inputs---\(\rho\), \(\theta\), and \(\psi\)---and outputs a \(Q\)-value for each of the five possible advisories.
The input space of each HCAS subsystem thus contains \(N_\mathrm{HCAS} = N_\rho \times N_\theta \times N_\psi = 32 \times 41 \times 41 = \num{53792}\) discrete training points, corresponding to the discretized values of \(\rho\), \(\theta\), and \(\psi\) listed in \cref{tab:HCAS_grid}.

\subsection{Vertical Collision Avoidance System (VCAS)}
VCAS issues vertical rate advisories to the ownship to avoid an NMAC with an intruder aircraft.
As shown in \cref{fig:VCAS}, the encounter is described by the relative altitude of the intruder with respect to the ownship \(h\), the ownship vertical rate \(\dot{h}_\mathrm{own}\), the intruder vertical rate \(\dot{h}_\mathrm{int}\), the time to the closest point of approach \(\tau\), and the previous advisory \(s_\mathrm{adv}\).
The complete parameter ranges used in this work are listed in \cref{tab:VCAS_ODD} and correspond to the open-source implementation\footnote{Available at \url{https://github.com/sisl/VerticalCAS}.}~\cite{Julian2019}.
The nine possible advisories are Clear of Conflict (COC), Do Not Climb (DNC), Do Not Descend (DND), Descend at least \qty{1500}{\foot\per\minute} (DES1500), Climb at least \qty{1500}{\foot\per\minute} (CL1500), Strengthen Descent to at least \qty{1500}{\foot\per\minute} (SDES1500), Strengthen Climb to at least \qty{1500}{\foot\per\minute} (SCL1500), Strengthen Descent to at least \qty{2500}{\foot\per\minute} (SDES2500), and Strengthen Climb to at least \qty{2500}{\foot\per\minute} (SCL2500), corresponding to \(s_\mathrm{adv} \in \{0, 1, 2, \dots, 8\}\)~\cite{Julian2019a}.

Similar to HCAS, VCAS is decomposed into an ensemble of nine individual neural networks, one for each possible previous advisory \(s_\mathrm{adv}\)~\cite{Julian2019, Julian2016}.
Each subsystem takes the four remaining state variables---\(h\), \(\dot{h}_\mathrm{own}\), \(\dot{h}_\mathrm{int}\), and \(\tau\)---as inputs and outputs a \(Q\)-value for each of the nine possible advisories.
The input space of each VCAS subsystem thus contains \(N_\mathrm{VCAS} = N_h \times N_{\dot{h}_\mathrm{own}} \times N_{\dot{h}_\mathrm{int}} \times N_\tau = 65 \times 39 \times 39 \times 41 = \num{4053465}\) discrete training points, corresponding to the discretized values of \(h\), \(\dot{h}_\mathrm{own}\), \(\dot{h}_\mathrm{int}\), and \(\tau{}\) listed in \cref{tab:VCAS_grid}, approximately two orders of magnitude more than each HCAS subsystem.

\subsection{Advisory Selection}
For both HCAS and VCAS, the advisory issued at runtime is determined by selecting the action with the highest \(Q\)-value for the current state, i.e.,
\begin{equation}
    s_\mathrm{adv}^* = \arg\max_{a} Q(s, a)\text{,}
\end{equation}
where \(s\) is the current state and \(a\) ranges over all admissible advisories given the previous advisory \(s_\mathrm{adv}\)~\cite{Julian2016}.
Inadmissible transitions---issuing a climb advisory immediately following a strong descent advisory---are excluded to ensure advisory consistency and reduce pilot workload~\cite{ED-256, Julian2019}.
In the neural network-based implementation, the network replaces the original lookup table: given the current state, the network outputs estimated \(Q\)-values for all advisories, and the one with the highest estimated value is selected~\cite{Julian2019a, Julian2016}.
Provided the neural network accurately represents the underlying MDP, the resulting advisory sequence is equivalent to that of the original table.

\section{Development Process}\label{sec:developmentprocess}
\begin{figure}[htb]
    \centering
    \includegraphics[width=\linewidth]{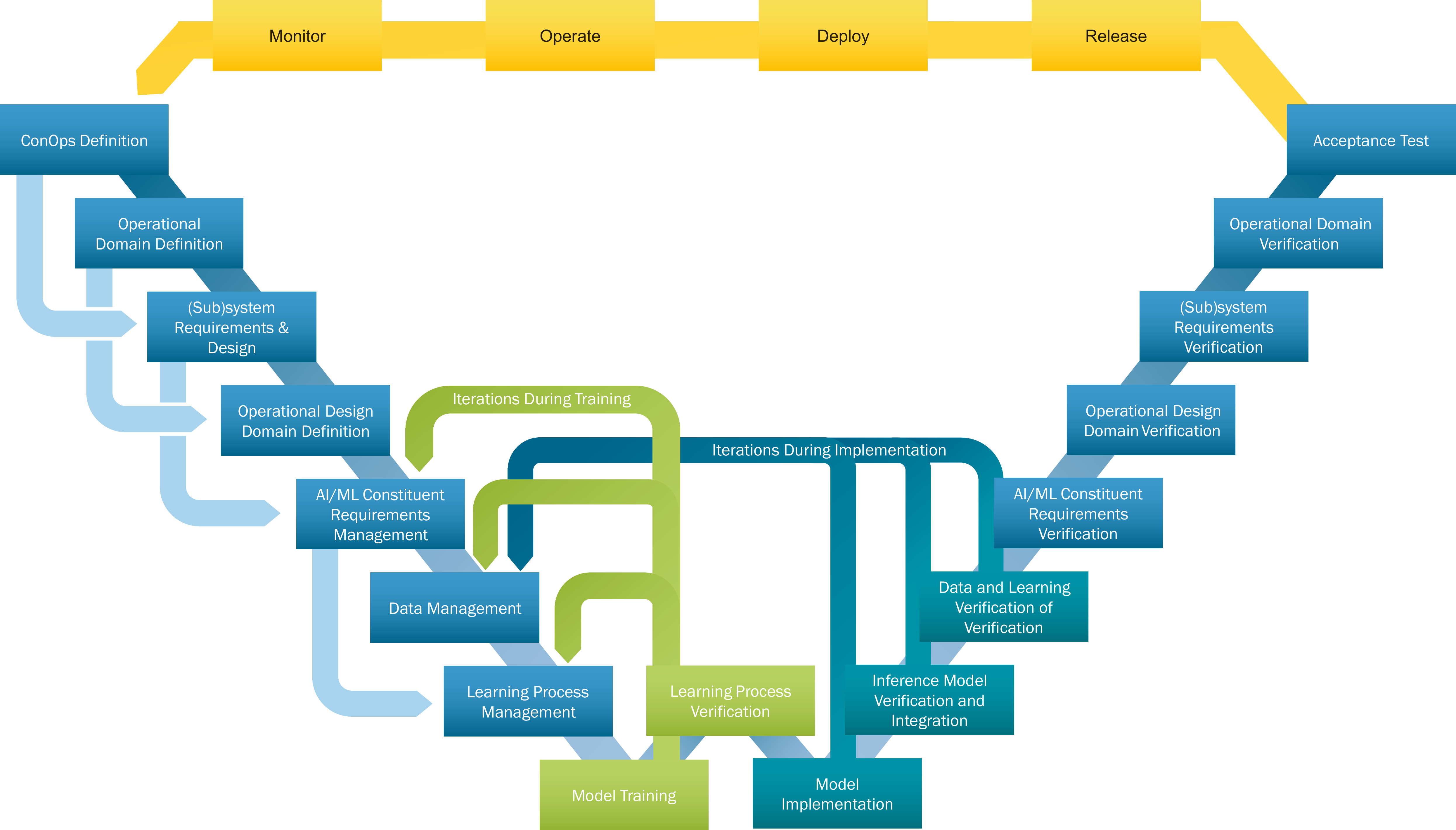}
    \caption{The extended W-shaped process~\cite{Christensen2025}, an extension of EASA's W-shaped process~\cite{EUASA2024}.}
    \label{fig:ExtWShapedProcess}
\end{figure}

As previously mentioned, EASA, aware of the trend towards AI-based applications in aviation, issued a concept paper detailing the steps required for future deployment of AI-based applications~\cite{EUASA2024}\footnote{All objective references in this work are to the published Issue~02~\cite{EUASA2024}. During the preparation of this paper, EASA released the Proposed Issue~03~\cite{EUASA2026}, which generalizes the scope from machine learning to artificial intelligence in general and renumbers the objectives under a new \emph{Technology-Phase-Number} convention. As Proposed Issue~03 is still under open consultation and thus subject to change, Issue~02 is retained as the primary reference; the implications for Safety Nets are discussed in \cref{sec:Certification}.}.
In their concept paper, EASA proposes the W-shaped development process.
Compared to the normally used V-model, it is augmented by dedicated steps concerning the model training and learning process verification.
However, previous research indicates that this is not enough to develop an AI-based system safely.
Instead, a more agile DevOps-based approach appears to be favorable~\cite{Beck2001, Lwakatare2020, Hubbs2023, Babu2024}.
Thus, the extended W-shaped process, see \cref{fig:ExtWShapedProcess}, has been proposed~\cite{Christensen2025}.
This process forms the basis of the development process for the AI/ML constituent in this work.

Accordingly, the first step is to define a Concept of Operations (ConOps) for the use case~\cite{Stefani2023}.
This has already been done by prior works~\cite{Werner2026, Stefani2023}.
For this work, the established ConOps is adopted directly: the system---HCAS and VCAS---is designed to provide pilots with last-resort measures to prevent mid-air collisions by issuing both vertical and horizontal resolution advisories.
The system is designed to operate in European Class~C airspace, where both the ownship and all possible intruders are equipped with an Automatic Dependent Surveillance-Broadcast (ADS-B) system.
No coordination between the ownship and the intruder is assumed, and the pilot must first assess whether the issued advisory can be executed safely before doing so.
This ConOps aligns directly with the operational assumptions underlying ACAS~Xa and ACAS~Xu as defined in the applicable standards~\cite{ED-256, ED-275, DO-385, DO-386}.

Next, the Operational Domain (OD) and ODD have to be defined.
Again, this has already been conducted by prior works to different levels of detail~\cite{Werner2026, Stefani2023, MLEAPConsortium2024}.
While the MLEAP report~\cite{MLEAPConsortium2024} explores an OD and ODD for ACAS~Xu/HCAS, both aligned with the ML input parameters of the ML inference---the parameters listed in \cref{tab:HCAS_ODD}---other works produced an OD and ODD more in line with concepts from automotive~\cite{Werner2026}.
This OD and ODD combination includes actual environmental and time-of-day constraints, such as weather conditions, but also the pilot's reaction time.
Nevertheless, for the context of this work, these ODDs are considered equivalent in their relevant aspects~\cite{Christensen2026}, and the MLEAP approach will be adopted.
This approach has the additional advantage of directly enabling the derivation of the OD and ODD from the underlying standards for ACAS~Xa and ACAS~Xu~\cite{ED-256, ED-275, DO-385, DO-386}, which explicitly state the required parameter ranges.

The parameter ranges for both HCAS and VCAS, listed in \cref{tab:HCAS_ODD,tab:VCAS_ODD}, can be derived from these standards.
For both systems, the maximum aircraft ground speed is bounded at \qty{600}{\knot}~\cite{ED-275, DO-386}.
For VCAS, vertical rate ranges of \([\qty{-100}{\foot\per\second}, \qty{100}{\foot\per\second}]\) for both ownship and intruder listed in \cref{tab:VCAS_ODD} are instead derived from the bounds of the open-source VerticalCAS implementation~\cite{Julian2019}, which are consistent with the aircraft performance limits referenced in the applicable standards~\cite{ED-256, ED-275, DO-385, DO-386}.
For HCAS, in higher altitude airspace with aircraft speeds up to \qty{600}{\knot}, the expected relative closing speed is no greater than \qty{566}{\knot} for intruders approaching from the side and no greater than \qty{400}{\knot} for intruders approaching from the rear~\cite{ED-256, DO-385}.
Furthermore, the discrete values of \(\tau\), the time to closest point of approach (CPA), used to partition the 40-network HCAS ensemble can be derived from Table~2-23 in \S{}2.2.4.6.4.2.3.1.10 of DO-386~\cite{DO-386}, where \(\tau\) was calculated at projection times \(\tau \in \{\qty{10}{\second}, \qty{20}{\second}, \qty{30}{\second}, \qty{40}{\second}, \qty{60}{\second}\}\) for initial ranges \(r_0 \in [\qty{3}{\nauticalmile}, \qty{30}{\nauticalmile}]\) and closing speeds \(v_0 \in [\qty{-1200}{\knot}, \qty{+1200}{\knot}]\)~\cite{DO-386}.
Here, the parameter ranges listed in \cref{tab:HCAS_ODD,tab:VCAS_ODD} represent a subset of the full operational parameter ranges defined in the standards, reflecting the proof-of-concept nature of HCAS and VCAS as open-source approximations of ACAS~Xa and ACAS~Xu~\cite{Julian2019}.
A detailed comparison of the use-case parameter ranges and resolution advisory sets against the full ACAS~Xa/Xu specifications is provided in \cref{tab:spec-coverage}, which quantifies the dimensions that are reduced, collapsed, or omitted relative to ED-256/DO-385 and ED-275/DO-386.
\begin{table}[htb]
    \caption{Parameter ranges for the HCAS use case. The ranges are slightly different than the original HCAS implementation~\cite{Julian2019} but match the corresponding GitHub release at \url{https://github.com/sisl/HorizontalCAS}.}
    \label{tab:HCAS_ODD}
    \centering
    \begin{tabular}{lll}
        \toprule
        Variable           & Description       & Range                                 \\
        \midrule
        \(\rho\)           & Range to intruder & [\qty{0}{\foot}, \qty{56000}{\foot}]  \\
        \(\theta\)         & Bearing angle     & [\ang{-180}, \ang{180}]               \\
        \(\psi\)           & Relative heading  & [\ang{-180}, \ang{180}]               \\
        \(v_\mathrm{own}\) & Ownship speed     & \qty{200}{\foot\per\second}           \\
        \(v_\mathrm{int}\) & Intruder speed    & \qty{200}{\foot\per\second}           \\
        \(\tau\)           & Time to CPA       & [\qty{0}{\second}, \qty{60}{\second}] \\
        \(s_\mathrm{adv}\) & Previous advisory & \{0, 1, 2, 3, 4\}                     \\
        \bottomrule
    \end{tabular}
\end{table}
\begin{table}[htb]
    \caption{Parameter ranges for the VCAS use case. The ranges are slightly different than the original VCAS implementation~\cite{Julian2019} but match the corresponding GitHub release at \url{https://github.com/sisl/VerticalCAS}.}
    \label{tab:VCAS_ODD}
    \centering
    \begin{tabular}{lll}
        \toprule
        Variable                 & Description          & Range                                                       \\
        \midrule
        \(h\)                    & Relative altitude    & [\qty{-8000}{\foot}, \qty{8000}{\foot}]                     \\
        \(\dot{h}_\mathrm{own}\) & Ownship vert.\ rate  & [\qty{-100}{\foot\per\second}, \qty{100}{\foot\per\second}] \\
        \(\dot{h}_\mathrm{int}\) & Intruder vert.\ rate & [\qty{-100}{\foot\per\second}, \qty{100}{\foot\per\second}] \\
        \(\tau{}\)               & Time to CPA          & [\qty{0}{\second}, \qty{40}{\second}]                       \\
        \(s_\mathrm{adv}\)       & Previous advisory    & \{0, 1, 2, 3, 4, 5, 6, 7, 8\}                                       \\
        \bottomrule
    \end{tabular}
\end{table}

The requirements for the AI/ML constituent are also derived directly from standards~\cite{DO-385, DO-386}.
The primary functional requirement for the AI/ML constituent is that its output must agree with the lookup table advisory for every possible state in the discretized input space, faithfully replicating the underlying MDP policy~\cite{Damour2021, Christensen2024}.
This is a strictly stronger requirement than a conventional accuracy target: it demands not merely high aggregate agreement, but a verifiably correct output for every valid input vector.
Satisfying this requirement through statistical testing alone is insufficient~\cite{EUASA2024}, as no finite test set can certify correctness across the complete input space.
Formal verification methods such as reachability analysis~\cite{ManzanasLopez2023, Katz2017} can provide stronger guarantees for bounded regions, but have been applied to HCAS and VCAS only for selected meta-properties rather than full policy agreement, and their computational cost scales unfavorably with input space size.
A Safety-by-Design approach is therefore required, wherein correctness is built into the system architecture rather than verified a posteriori.
Safety Nets fulfill this role by construction: because the lookup table stores the correct advisory for every input vector where the neural network is known to fail, the combined system is provably correct over the entire discrete input space.
This construction simultaneously satisfies a set of EASA learning assurance and implementation objectives through exhaustive verification against the ground truth, as enumerated in \cref{tab:safety-net-objectives}.

For data management, the training data are derived directly from the MDP lookup tables provided with the open-source implementations of HCAS and VCAS~\cite{Julian2019}.
These lookup tables constitute the formal ground truth against which both training and verification are performed.
Because the objective is to represent the complete, finite input space correctly---rather than to generalize from a sample to unseen data---the entire dataset serves simultaneously as the training, validation, and testing set.
This departs from the standard machine learning paradigm of disjoint data splits, but is fully consistent with prior work in this domain~\cite{Julian2019a, Damour2021}: the problem is not generalization, but minimally lossy compression of a known, discrete function.
From a learning assurance perspective, this design choice is intentional.
By evaluating the neural network against the ground truth at every discrete point in the ODD, the Safety Net construction process fulfills the verification objectives of~\cite{EUASA2024} listed in \cref{tab:safety-net-objectives} without requiring a separate post-training verification campaign.

\section{Performance Study}\label{sec:PerformanceStudy}
The hyperparameter study and Safety Net construction described in this and the following section instantiate the \emph{Data Management}, \emph{Learning Process Management}, and \emph{Model Training} phases of the extended W-shaped process (\cref{fig:ExtWShapedProcess}); the exhaustive verification sweep that produces each lookup table corresponds to its \emph{Learning Process Verification} and \emph{Inference Model Verification and Integration} phases.
HCAS and VCAS both use a neural network to store an estimated \(Q\)-value based on the current state.
Depending on the estimated value, ACAS~Xa/Xu, and by extension HCAS and VCAS, issue a resolution advisory to the pilots to avoid a near mid-air collision.
In the original implementation of HCAS and VCAS, each system was represented not by a single neural network but by an ensemble of neural networks to increase the accuracy while keeping the overall system as small as possible~\cite{Julian2016}.
Moreover, the neural networks were trained to learn all \(Q\)-values for each state across all possible RAs, even those that would not be selected.
If instead one-hot encoding is used, where the highest \(Q\)-value is set to one, and all other outputs are zero, thereby effectively turning the regression problem into a classification problem, performance generally improves as expected~\cite{Bishop2006, Golik2013}.
Since both HCAS and VCAS issue only the advisory with the highest \(Q\)-value (cf.\ \cref{sec:UseCase}), discarding the \(Q\)-values themselves does not alter the functional behavior of the system at the discrete grid points.
For the training of the neural networks in this work, Mean Squared Error Loss was used when attempting to learn the full \(Q\)-value vector---equal to previous research~\cite{Julian2019}---while Cross Entropy Loss was used for the one-hot encoded output vector, as is standard with classification problems.
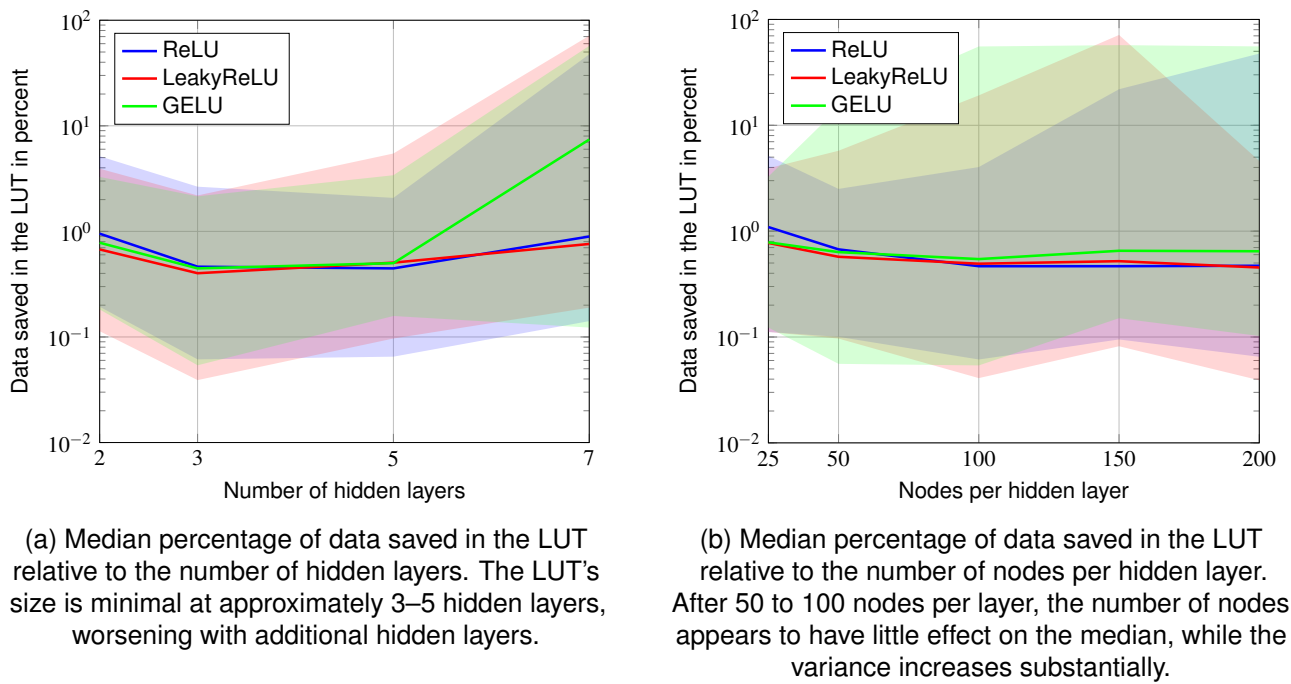
\begin{figure}[htb]
    \begin{subfigure}[t]{0.47\linewidth}
        \centering
        \resizebox{\linewidth}{!}{%
            \begin{tikzpicture}
    \pgfplotsset{
        every axis/.style={
            scale only axis,
            colormap/viridis,
            ymode=log,
            xlabel={Number of hidden layers},
            ylabel={Data saved in the LUT in percent},
            xmin=2,
            xmax=7,
            ymin=0.01,
            ymax=100,
            xtick={2, 3, 5, 7},
            ymajorgrids,
            xmajorgrids,
            legend columns=1,
            legend style={
                legend cell align=left,
                align=left,
                legend pos=north west,
            }
        },
    }
    
    \pgfdeclarelayer{background}
    \pgfdeclarelayer{foreground}
    \pgfsetlayers{background,main,foreground}
    
    \newcommand{\plotActivation}[2]{%
        \pgfplotstableread[col sep=comma] {./data/data_HCAS_#1_n_hidden_layers.csv} \datatable;
        
        \begin{pgfonlayer}{axis background}
            \addplot[name path=U#1, draw=none, forget plot, line width=1pt,
                table/x=n_hidden_layers, table/y=in_lut_max] table {\datatable};
            \addplot[name path=L#1, draw=none, forget plot, line width=1pt,
                table/x=n_hidden_layers, table/y=in_lut_min] table {\datatable};
            \addplot[color=#2!80, opacity=0.2, forget plot]
                fill between [of=U#1 and L#1];
        \end{pgfonlayer}
        
        \begin{pgfonlayer}{axis foreground}
            \addplot[color=#2, line width=1.25pt,
                table/x=n_hidden_layers, table/y=in_lut_median] table {\datatable};
        \end{pgfonlayer}
        
        \addlegendentry{#1};
    }
    
    \begin{axis}
        \plotActivation{ReLU}{blue}
        \plotActivation{LeakyReLU}{red}
        \plotActivation{GELU}{green}
    \end{axis}

\end{tikzpicture}%
        }%
        \caption{Median percentage of data saved in the LUT relative to the number of hidden layers. The LUT's size is minimal at approximately 3--5 hidden layers, worsening with additional hidden layers.}
        \label{fig:HCAS_performance_hidden_layers}
    \end{subfigure}
    \hfill
    \begin{subfigure}[t]{0.48\linewidth}
        \centering
        \resizebox{\linewidth}{!}{%
            \begin{tikzpicture}
    \pgfplotsset{
        every axis/.style={
            scale only axis,
            ymode=log,
            xlabel={Nodes per hidden layer},
            ylabel={Data saved in the LUT in percent},
            xmin=25,
            xmax=200,
            ymin=0.01,
            ymax=100,
            xtick={25, 50, 100, 150, 200},
            ymajorgrids,
            xmajorgrids,
            legend columns=1,
            legend style={
                legend cell align=left,
                align=left,
                legend pos=north west,
            }
        },
    }
    
    \pgfdeclarelayer{background}
    \pgfdeclarelayer{foreground}
    \pgfsetlayers{background,main,foreground}
    
    \newcommand{\plotActivation}[2]{%
        \pgfplotstableread[col sep=comma] {./data/data_HCAS_#1_layer_size.csv} \datatable;
        
        \begin{pgfonlayer}{axis background}
            \addplot[name path=U#1, draw=none, forget plot, line width=1pt,
                table/x=layer_size, table/y=in_lut_max] table {\datatable};
            \addplot[name path=L#1, draw=none, forget plot, line width=1pt,
                table/x=layer_size, table/y=in_lut_min] table {\datatable};
            \addplot[color=#2!80, opacity=0.2, forget plot]
                fill between [of=U#1 and L#1];
        \end{pgfonlayer}
        
        \begin{pgfonlayer}{axis foreground}
            \addplot[color=#2, line width=1.25pt,
                table/x=layer_size, table/y=in_lut_median] table {\datatable};
        \end{pgfonlayer}
        
        \addlegendentry{#1};
    }
    
    \begin{axis}
        \plotActivation{ReLU}{blue}
        \plotActivation{LeakyReLU}{red}
        \plotActivation{GELU}{green}
    \end{axis}

\end{tikzpicture}%
        }%
        \caption{Median percentage of data saved in the LUT relative to the number of nodes per hidden layer. After 50 to 100 nodes per layer, the number of nodes appears to have little effect on the median, while the variance increases substantially.}
        \label{fig:HCAS_performance_nodes_per_layer}
    \end{subfigure}
    \caption{Logarithmic plots displaying the impact of the number of hidden layers and nodes per hidden layer of the fully connected neural network on the retrieval rate for the three activation functions ReLU, LeakyReLU, and GELU. For the encoding, one-hot was used. The data are aggregated across all 40 HCAS subsystems. Lower percentages of data saved in the LUT correspond to smaller system sizes.}
    \label{fig:HCAS_performance}
\end{figure}
\begin{figure}[htb]
    \begin{subfigure}[t]{0.47\linewidth}
        \centering
        \resizebox{\linewidth}{!}{%
            \begin{tikzpicture}
    \pgfplotsset{
        every axis/.style={
            scale only axis,
            colormap/viridis,
            ymode=log,
            xlabel={Number of hidden layers},
            ylabel={Data saved in the LUT in percent},
            xmin=2,
            xmax=7,
            ymin=0.01,
            ymax=100,
            xtick={2, 3, 5, 7},
            ymajorgrids,
            xmajorgrids,
            legend columns=1,
            legend style={
                legend cell align=left,
                align=left,
                legend pos=north west,
            }
        },
    }
    
    \pgfdeclarelayer{background}
    \pgfdeclarelayer{foreground}
    \pgfsetlayers{background,main,foreground}
    
    \newcommand{\plotActivation}[2]{%
        \pgfplotstableread[col sep=comma] {./data/data_VCAS_#1_n_hidden_layers.csv} \datatable;
        
        \begin{pgfonlayer}{axis background}
            \addplot[name path=U#1, draw=none, forget plot, line width=1pt,
                table/x=n_hidden_layers, table/y=in_lut_max] table {\datatable};
            \addplot[name path=L#1, draw=none, forget plot, line width=1pt,
                table/x=n_hidden_layers, table/y=in_lut_min] table {\datatable};
            \addplot[color=#2!80, opacity=0.2, forget plot]
                fill between [of=U#1 and L#1];
        \end{pgfonlayer}
        
        \begin{pgfonlayer}{axis foreground}
            \addplot[color=#2, line width=1.25pt,
                table/x=n_hidden_layers, table/y=in_lut_median] table {\datatable};
        \end{pgfonlayer}
        
        \addlegendentry{#1};
    }
    
    \begin{axis}
        \plotActivation{ReLU}{blue}
        \plotActivation{LeakyReLU}{red}
        \plotActivation{GELU}{green}
    \end{axis}

\end{tikzpicture}%
        }%
        \caption{Median percentage of data saved in the LUT relative to the number of hidden layers. The LUT's size is minimal at 3--5 hidden layers, depending on the activation function, worsening with additional hidden layers.}
        \label{fig:VCAS_performance_hidden_layers}
    \end{subfigure}
    \hfill
    \begin{subfigure}[t]{0.48\linewidth}
        \centering
        \resizebox{\linewidth}{!}{%
            \begin{tikzpicture}
    \pgfplotsset{
        every axis/.style={
            scale only axis,
            ymode=log,
            xlabel={Nodes per hidden layer},
            ylabel={Data saved in the LUT in percent},
            xmin=25,
            xmax=200,
            ymin=0.01,
            ymax=100,
            xtick={25, 50, 100, 150, 200},
            ymajorgrids,
            xmajorgrids,
            legend columns=1,
            legend style={
                legend cell align=left,
                align=left,
                legend pos=north west,
            }
        },
    }
    
    \pgfdeclarelayer{background}
    \pgfdeclarelayer{foreground}
    \pgfsetlayers{background,main,foreground}
    
    \newcommand{\plotActivation}[2]{%
        \pgfplotstableread[col sep=comma] {./data/data_VCAS_#1_layer_size.csv} \datatable;
        
        \begin{pgfonlayer}{axis background}
            \addplot[name path=U#1, draw=none, forget plot, line width=1pt,
                table/x=layer_size, table/y=in_lut_max] table {\datatable};
            \addplot[name path=L#1, draw=none, forget plot, line width=1pt,
                table/x=layer_size, table/y=in_lut_min] table {\datatable};
            \addplot[color=#2!80, opacity=0.2, forget plot]
                fill between [of=U#1 and L#1];
        \end{pgfonlayer}
        
        \begin{pgfonlayer}{axis foreground}
            \addplot[color=#2, line width=1.25pt,
                table/x=layer_size, table/y=in_lut_median] table {\datatable};
        \end{pgfonlayer}
        
        \addlegendentry{#1};
    }
    
    \begin{axis}
        \plotActivation{ReLU}{blue}
        \plotActivation{LeakyReLU}{red}
        \plotActivation{GELU}{green}
    \end{axis}

\end{tikzpicture}%
        }%
        \caption{Median percentage of data saved in the LUT relative to the number of nodes per hidden layer. After 50 to 100 nodes per layer, the number of nodes appears to have little effect on the median, while the variance increases substantially.}
        \label{fig:VCAS_performance_nodes_per_layer}
    \end{subfigure}
    \caption{Logarithmic plots displaying the impact of the number of hidden layers and nodes per hidden layer of the fully connected neural network on the retrieval rate for the three activation functions ReLU, LeakyReLU, and GELU. For the encoding, one-hot was used. The data are aggregated across all 9 VCAS subsystems. Lower percentages of data saved in the LUT correspond to smaller system sizes.}
    \label{fig:VCAS_performance}
\end{figure}
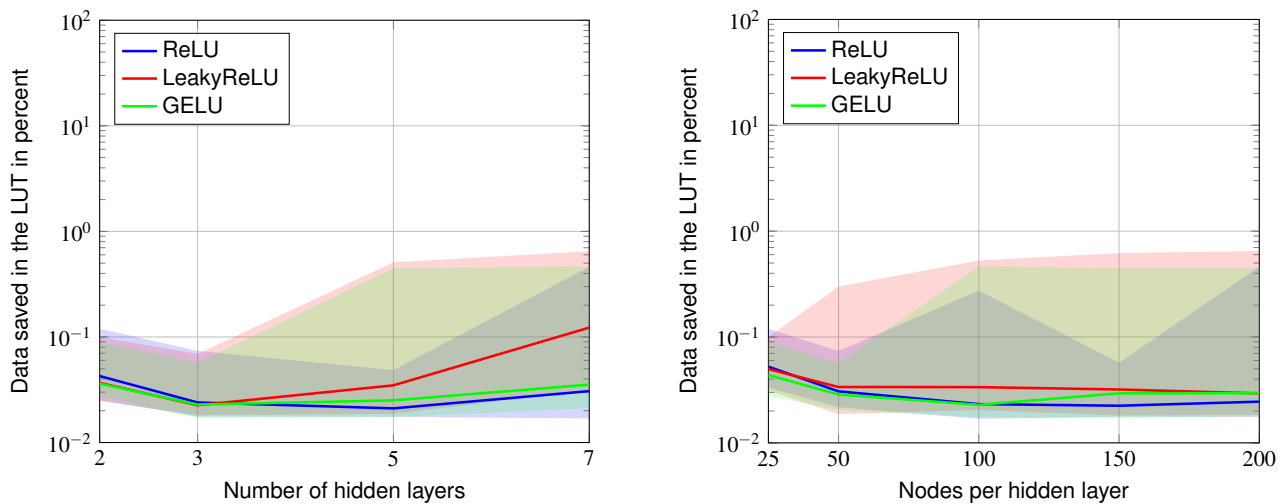
\begin{figure}[htb]
    \begin{subfigure}[t]{0.47\linewidth}
        \centering
        \resizebox{\linewidth}{!}{%
            \begin{tikzpicture}
    \pgfplotsset{
        every axis/.style={
            scale only axis,
            colormap/viridis,
            ymode=log,
            xlabel={Number of hidden layers},
            ylabel={Data saved in the LUT in percent},
            xmin=2,
            xmax=7,
            ymin=0.01,
            ymax=100,
            xtick={2, 3, 5, 7},
            ymajorgrids,
            xmajorgrids,
            legend columns=1,
            legend style={
                legend cell align=left,
                align=left,
                legend pos=north west,
            }
        },
    }
    
    \pgfdeclarelayer{background}
    \pgfdeclarelayer{foreground}
    \pgfsetlayers{background,main,foreground}
    
    \newcommand{\plotActivation}[3]{%
        \pgfplotstableread[col sep=comma] {./data/data_VCAS_ReLU_n_hidden_layers#1.csv} \datatable;
        
        \begin{pgfonlayer}{axis background}
            \addplot[name path=U#1, draw=none, forget plot, line width=1pt,
                table/x=n_hidden_layers, table/y=in_lut_max] table {\datatable};
            \addplot[name path=L#1, draw=none, forget plot, line width=1pt,
                table/x=n_hidden_layers, table/y=in_lut_min] table {\datatable};
            \addplot[color=#2!80, opacity=0.2, forget plot]
                fill between [of=U#1 and L#1];
        \end{pgfonlayer}
        
        \begin{pgfonlayer}{axis foreground}
            \addplot[color=#2, line width=1.25pt,
                table/x=n_hidden_layers, table/y=in_lut_median] table {\datatable};
        \end{pgfonlayer}
        
        \addlegendentry{#3};
    }
    
    \begin{axis}
        \plotActivation{}{blue}{One hot}
        \plotActivation{-no-one-hot}{red}{Target encoding}
    \end{axis}

\end{tikzpicture}%
        }%
        \caption{Median percentage of data saved in the LUT relative to the number of hidden layers.}
        \label{fig:VCAS_performance_hidden_layers_one_hot}
    \end{subfigure}
    \hfill
    \begin{subfigure}[t]{0.48\linewidth}
        \centering
        \resizebox{\linewidth}{!}{%
            \begin{tikzpicture}
    \pgfplotsset{
        every axis/.style={
            scale only axis,
            ymode=log,
            xlabel={Nodes per hidden layer},
            ylabel={Data saved in the LUT in percent},
            xmin=25,
            xmax=200,
            ymin=0.01,
            ymax=100,
            xtick={25, 50, 100, 150, 200},
            ymajorgrids,
            xmajorgrids,
            legend columns=1,
            legend style={
                legend cell align=left,
                align=left,
                legend pos=north west,
            }
        },
    }
    
    \pgfdeclarelayer{background}
    \pgfdeclarelayer{foreground}
    \pgfsetlayers{background,main,foreground}
    
    \newcommand{\plotActivation}[3]{%
        \pgfplotstableread[col sep=comma] {./data/data_VCAS_ReLU_layer_size#1.csv} \datatable;
        
        \begin{pgfonlayer}{axis background}
            \addplot[name path=U#1, draw=none, forget plot, line width=1pt,
                table/x=layer_size, table/y=in_lut_max] table {\datatable};
            \addplot[name path=L#1, draw=none, forget plot, line width=1pt,
                table/x=layer_size, table/y=in_lut_min] table {\datatable};
            \addplot[color=#2!80, opacity=0.2, forget plot]
                fill between [of=U#1 and L#1];
        \end{pgfonlayer}
        
        \begin{pgfonlayer}{axis foreground}
            \addplot[color=#2, line width=1.25pt,
                table/x=layer_size, table/y=in_lut_median] table {\datatable};
        \end{pgfonlayer}
        
        \addlegendentry{#3};
    }
    
    \begin{axis}
        \plotActivation{}{blue}{One hot}
        \plotActivation{-no-one-hot}{red}{Target encoding}
    \end{axis}

\end{tikzpicture}%
        }%
        \caption{Median percentage of data saved in the LUT relative to the number of nodes per hidden layer.}
        \label{fig:VCAS_performance_nodes_per_layer_one_hot}
    \end{subfigure}
    \caption{Logarithmic plots displaying the impact of the encoding---target encoding or one-hot---on the retrieval rate of the fully connected neural networks. Here, with a ReLU activation function and compartmentalized into a number of hidden layers and nodes per hidden layer. Lower percentages of data saved in the LUT correspond to smaller system sizes. For smaller neural networks, one-hot consistently outperforms target encoding on the median retrieval rate by at least one order of magnitude. The data are aggregated across all 9 VCAS subsystems.}
    \label{fig:VCAS_performance_one_hot}
\end{figure}
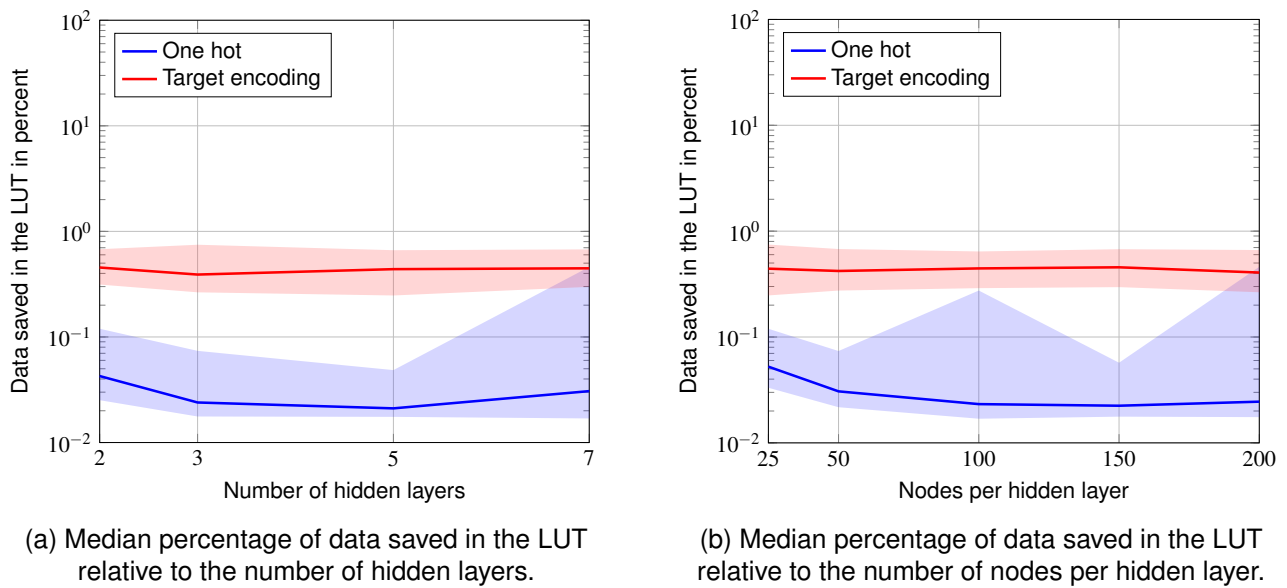

Because the Safety Net's overall size is determined jointly by the neural network architecture and the resulting LUT, a systematic hyperparameter study is required to identify the design that minimizes the total memory requirements.
A smaller LUT---achieved by a higher neural network \emph{retrieval rate}, defined as the fraction of input vectors the network classifies correctly so that no LUT lookup is required (equivalently, \qty{100}{\percent} minus the percentage of the input space stored in the LUT)---directly reduces system size, provided the network itself does not grow disproportionately large.
Preliminary studies identified that the ranges between 2 and 5 for the number of hidden layers and 50 to 150 for the number of nodes per hidden layer were the most promising to maximize the retrieval rate.
The three activation functions, ReLU, LeakyReLU, and GELU, were chosen based on current recommendations from prior research~\cite{Dubey2019, Maas2013, Xu2015}.
This leads to the following ranges for the hyperparameters: for the activation function ReLU, LeakyReLU, and GELU; for the number of hidden layers 2, 3, 5, and 7; for the number of nodes per hidden layer 25, 50, 100, 150, and 200; and for the encoding either target encoding---representing the full correct \(Q\)-value vector---or one-hot encoding.
Furthermore, for every combination of activation function, number of hidden layers, nodes per hidden layer, and encoding, training was conducted five times per neural network to reduce random influences.
Training for each neural network was performed for up to \num{10000} epochs, with early stopping after \num{1000} consecutive epochs without improvement---also called \emph{patience}---where one epoch is a complete pass through the training dataset.
Given the nature of the problem---exact memorization of a finite, known function rather than generalization to unseen samples---which requires a correct representation of the entire problem space, the entire dataset was used for training, validation, and testing; there is no held-out distribution to overfit to.
The early-stopping improvement criterion is therefore evaluated as the training loss---equivalently, the agreement rate against the complete discrete ground truth---while the patience guards only against wasted computation once convergence stalls, not against a loss of generalization.
Thus, the trained system has learned from and been tested on every possible scenario, ensuring accurate outputs across real-world applications.
This is consistent with prior work, which likewise trains and evaluates on the full dataset and notes that here overfitting is in fact encouraged~\cite{Julian2019a, Damour2021}.
After successfully training the neural network, the corresponding lookup table was created.

The results of the parameter studies are shown in \cref{fig:HCAS_performance} for HCAS and \cref{fig:VCAS_performance} for VCAS, with the percentage of the input space stored in the LUT plotted against the number of hidden layers and the number of nodes per hidden layer, separated by activation function, using one-hot encoding throughout.
A comparison between one-hot and target encoding is shown in \cref{fig:VCAS_performance_one_hot}.
In all plots, the median is shown with shading indicating the full range across the five training runs per configuration.
Because the neural network's memory footprint depends only on its architecture and not on the learned weight values, any reduction in the percentage of inputs stored in the LUT translates directly to a smaller combined system.
The median percentage of inputs stored in the LUT for VCAS is consistently one to two orders of magnitude lower than for HCAS.
Two structural differences between the two systems explain this gap.
First, each VCAS subsystem contains \num{4053465} training points, whereas each HCAS subsystem contains only \num{53792}, giving the VCAS subsystem almost two orders of magnitude more information per epoch, matching the performance advantage of VCAS.
Second, the HCAS input space includes two angular variables---the bearing angle \(\theta\) and the relative heading \(\psi\)---each spanning \ang{-180} to \ang{180}.
The periodicity of these variables likely introduces discontinuities at the wrap-around boundary that fully connected networks with standard weight initialization struggle to represent faithfully, resulting in systematically more absolute errors in HCAS subsystems and therefore a larger LUT~\cite{Rahaman2019, Sitzmann2020}.
A natural remedy, left to future work, is to encode the periodic angular inputs \(\theta{}\) and \(\psi{}\) by their sine and cosine components, removing the artificial discontinuity at the \ang{\pm180} wrap-around boundary that fully connected networks with standard initialization represent poorly~\cite{Rahaman2019, Sitzmann2020}; this would be expected to reduce the HCAS LUT even further.
VCAS inputs, by contrast, are all bounded continuous quantities (relative altitude, vertical rates, time to CPA) whose policy structure is largely monotonic, presenting a considerably more tractable regression surface.
First, ReLU persistently yields a lower median LUT percentage than LeakyReLU or GELU, despite prior literature suggesting the opposite for general classification tasks~\cite{Dubey2019, Maas2013, Xu2015}.
This is attributable to the discrete, bounded nature of the input space: ReLU's hard zeroing produces sparser, more piecewise-constant activation patterns~\cite{Montufar2014} that align well with the step-like decision boundaries of the underlying MDP policy, whereas the non-zero negative slope of LeakyReLU likely introduces unnecessary gradient signal in inactive regions.
GELU's smooth nonlinearity, beneficial for large-scale language and vision tasks, similarly offers no advantage here and incurs higher variance across runs~\cite{Hendrycks2016, Devlin2019}.
Second, increasing the number of nodes per hidden layer beyond 50 to 100 yields diminishing returns on the median LUT percentage while substantially increasing variance across training runs.
This indicates that the policy's representational complexity is saturated at moderate widths; additional capacity is not exploited consistently, suggesting multiple equivalent local minima in the loss landscape at higher widths.
Third, the optimal number of hidden layers lies between 3 and 5.
Shallower networks lack the depth to capture the piecewise geometry of the MDP policy, while networks deeper than 5 layers show a deteriorating median, likely because the optimization landscape becomes harder to navigate.

Based on the above, a Safety Net was created for both HCAS and VCAS using the best-performing configurations identified in the study.
The open-source implementations, together with reproducibility scripts, are released alongside this paper.

\section{Safety Net}\label{sec:SafetyNet}
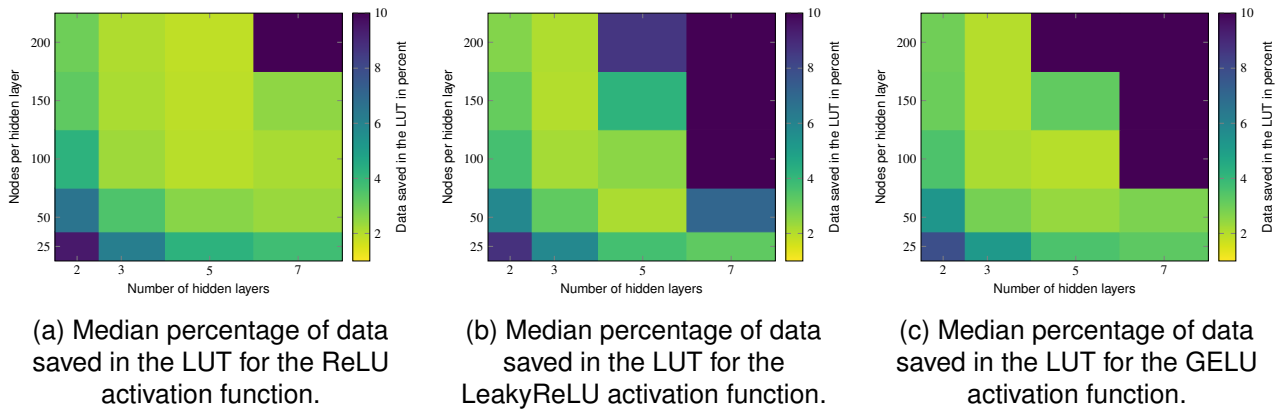
\begin{figure}[htb]
    \begin{subfigure}[t]{0.32\linewidth}
        \centering
        \resizebox{\linewidth}{!}{%
            \begin{tikzpicture}
    \pgfplotsset{
        every axis/.style={
            scale only axis,
            colorbar,
            colorbar style={
                ylabel=Data saved in the LUT in percent,
            },
            colormap={reverse viridis}{
                indices of colormap={
                    \pgfplotscolormaplastindexof{viridis},...,0 of viridis}
            },
            view={0}{90},
            xlabel={Number of hidden layers},
            ylabel={Nodes per hidden layer},
            zlabel={Data saved in the LUT},
            xtick={2, 3, 5, 7},
            ytick={25, 50, 100, 150, 200},
            enlargelimits=false,
            axis on top,
            point meta min=1,
            point meta max=10,
        },
    }
    
    \newcommand{\plotHeatMap}[3]{%
        \addplot[
        matrix plot*,
        point meta=explicit,
        ]
        table[
            col sep=space,
            x index=0,
            y index=1,
            meta expr=\thisrowno{2}*100
        ] {./data/matrix_#1_#2_#3.dat};
    }
    
    \begin{axis}
        \plotHeatMap{VCAS}{ReLU}{median}
    \end{axis}

\end{tikzpicture}%
        }%
        \caption{Median percentage of data saved in the LUT for the ReLU activation function.}
        \label{fig:VCAS_performance_heatmap_ReLU}
    \end{subfigure}
    \hfill
    \begin{subfigure}[t]{0.32\linewidth}
        \centering
        \resizebox{\linewidth}{!}{%
            \begin{tikzpicture}
    \pgfplotsset{
        every axis/.style={
            scale only axis,
            colorbar,
            colorbar style={
                ylabel=Data saved in the LUT in percent,
            },
            colormap={reverse viridis}{
                indices of colormap={
                    \pgfplotscolormaplastindexof{viridis},...,0 of viridis}
            },
            view={0}{90},
            xlabel={Number of hidden layers},
            ylabel={Nodes per hidden layer},
            zlabel={Data saved in the LUT},
            xtick={2, 3, 5, 7},
            ytick={25, 50, 100, 150, 200},
            enlargelimits=false,
            axis on top,
            point meta min=1,
            point meta max=10,
        },
    }
    
    \newcommand{\plotHeatMap}[3]{%
        \addplot[
        matrix plot*,
        point meta=explicit,
        ]
        table[
            col sep=space,
            x index=0,
            y index=1,
            meta expr=\thisrowno{2}*100
        ] {./data/matrix_#1_#2_#3.dat};
    }
    
    \begin{axis}
        \plotHeatMap{VCAS}{LeakyReLU}{median}
    \end{axis}

\end{tikzpicture}%
        }%
        \caption{Median percentage of data saved in the LUT for the LeakyReLU activation function.}
        \label{fig:VCAS_performance_heatmap_LeakyReLU}
    \end{subfigure}
    \hfill
    \begin{subfigure}[t]{0.32\linewidth}
        \centering
        \resizebox{\linewidth}{!}{%
            \begin{tikzpicture}
    \pgfplotsset{
        every axis/.style={
            scale only axis,
            colorbar,
            colorbar style={
                ylabel=Data saved in the LUT in percent,
            },
            colormap={reverse viridis}{
                indices of colormap={
                    \pgfplotscolormaplastindexof{viridis},...,0 of viridis}
            },
            view={0}{90},
            xlabel={Number of hidden layers},
            ylabel={Nodes per hidden layer},
            zlabel={Data saved in the LUT},
            xtick={2, 3, 5, 7},
            ytick={25, 50, 100, 150, 200},
            enlargelimits=false,
            axis on top,
            point meta min=1,
            point meta max=10,
        },
    }
    
    \newcommand{\plotHeatMap}[3]{%
        \addplot[
        matrix plot*,
        point meta=explicit,
        ]
        table[
            col sep=space,
            x index=0,
            y index=1,
            meta expr=\thisrowno{2}*100
        ] {./data/matrix_#1_#2_#3.dat};
    }
    
    \begin{axis}
        \plotHeatMap{VCAS}{GELU}{median}
    \end{axis}

\end{tikzpicture}%
        }%
        \caption{Median percentage of data saved in the LUT for the GELU activation function.}
        \label{fig:VCAS_performance_heatmap_GELU}
    \end{subfigure}
    \hfill
    \caption{Heatmaps displaying the impact of the number of hidden layers and nodes per hidden layer, split into the three activation functions---ReLU, LeakyReLU, and GELU. Compared to LeakyReLU and GELU, ReLU appears to be more forgiving in choosing the right combination of hidden layers and nodes per hidden layer. The data are aggregated across all 9 VCAS subsystems.}
    \label{fig:VCAS_performance_heatmap}
\end{figure}

Given the results of \cref{sec:PerformanceStudy}, followed by a joint hyperparameter sensitivity analysis, shown in \cref{fig:VCAS_performance_heatmap} for VCAS, the final Safety Nets for both HCAS and VCAS were assembled using the best-performing configurations identified in the study.
For both systems, ReLU was selected as the activation function, despite prior literature suggesting a preference for LeakyReLU in general classification tasks~\cite{Dubey2019, Maas2013, Xu2015}.
As discussed in \cref{sec:PerformanceStudy}, this choice is well-motivated by the discrete, bounded nature of the MDP policy underlying both HCAS and VCAS, for which ReLU's piecewise-constant activation patterns are particularly well-suited.
One-hot encoding was adopted for both systems, as it consistently outperformed the target encoding by at least one order of magnitude on the median LUT percentage, particularly for smaller network architectures.
For HCAS, an ensemble of networks with 3 hidden layers and 100 nodes per hidden layer was selected, while for VCAS, 5 hidden layers and 100 nodes per hidden layer were selected, both with training performed for up to \num{10000} epochs and early stopping after \num{1000} consecutive epochs without improvement.
Consequently, this configuration falls right into the identified optimal range of 3 to 5 hidden layers and approximately 50 to 100 nodes per layer, beyond which the median LUT percentage does not improve while variance increases substantially and the neural network's own memory footprint grows, increasing the combined system size.
In total, on a workstation equipped with an NVIDIA GeForce RTX~4090 GPU and a 13th-generation Intel Core i9-13900K CPU with \qty{64}{\gibi\byte} of RAM, the training took around \qty{77}{\hour} for HCAS and \qty{269}{\hour} for VCAS.
The resulting aggregate system sizes are summarized in \cref{tab:SafetyNetSizes}.
\begin{table}[htb]
    \caption{%
        Aggregate sizes of the final Safety Nets for HCAS and VCAS, based on a \texttt{scipy.spatial} k-d tree implementation of the LUT\@.
        All 40 HCAS and 9 VCAS subsystems are included.%
    }
    \label{tab:SafetyNetSizes}
    \centering
    \begin{tabular}{lrrrr}
        \toprule
        System & Neural network total   & LUT total                & Combined                 & Avg.\ neural network coverage \\
        \midrule
        HCAS   & \qty{3.22}{\mebi\byte} & \qty{1.54}{\mebi\byte}   & \qty{4.76}{\mebi\byte}   & \qty{99.65}{\percent}         \\
        VCAS   & \qty{1.44}{\mebi\byte} & \qty{221.40}{\mebi\byte} & \qty{222.83}{\mebi\byte} & \qty{97.77}{\percent}         \\
        \bottomrule
    \end{tabular}
\end{table}
The average neural network coverage across all 40 HCAS subsystems of \qty{99.65}{\percent} exceeds the median results of \cref{sec:PerformanceStudy}, as the final ensemble retains the best-performing of the five training runs for each subsystem, whereas \cref{fig:HCAS_performance} reports medians across runs aggregated over all subsystems.
Compared to a purely lookup table-based implementation of ACAS~X, which requires at least \qty{4}{\gibi\byte} of memory, the combined Safety Nets occupy \qty{4.76}{\mebi\byte} for HCAS and \qty{222.83}{\mebi\byte} for VCAS---a reduction of almost three orders of magnitude and slightly more than one order of magnitude, respectively.
Thus, both systems would fit within the memory budget of current avionics hardware, rendering deployment feasible.
The LUT sizes reported here use a \texttt{scipy.spatial} k-d tree implementation~\cite{Bentley1975, Virtanen2020}, which reduces the LUT footprint by roughly a factor of three relative to a na\"ive Python dictionary (from \qty{5.27}{\mebi\byte} to \qty{1.54}{\mebi\byte} for HCAS and from \qty{677.40}{\mebi\byte} to \qty{221.40}{\mebi\byte} for VCAS).

Beyond storage, runtime performance is critical for deployment on avionics hardware.
The ACAS~X surveillance-and-resolution cycle operates at approximately \qty{1}{\hertz}, i.e., one advisory update per second, as specified in the ED-256/DO-385 and ED-275/DO-386, for ACAS~Xa and ACAS~Xu, respectively~\cite{ED-256, ED-275, DO-385, DO-386}; the inference time of the AI/ML constituent must therefore remain well below \qty{1}{\second}.
The combined Safety Nets were benchmarked on the same workstation as was used for training the Safety Net, averaging over \num{5000} samples per subsystem.
On the GPU, HCAS achieves a mean inference time of \qty{33.2 +- 1.5}{\micro\second} per sample and VCAS \qty{57.3 +- 12.7}{\micro\second} per sample.
On the CPU alone, both systems are even faster, at \qty{20.2 +- 4.3}{\micro\second} for HCAS and \qty{46.4 +- 13.0}{\micro\second} for VCAS, owing to the modest network sizes that avoid host-device transfer overhead.
These sub-millisecond inference times sit roughly four orders of magnitude below the \qty{1}{\second} update budget, confirming ample runtime feasibility of the approach.

In line with the Safety Net concept~\cite{Christensen2024}, inference is defined only at the discrete input vectors enumerated in the manifest (cf.\ \cref{tab:HCAS_grid,tab:VCAS_grid}): continuous runtime states are mapped to the closest available discrete input vector before the check module is consulted.
The \qty{100}{\percent} correctness guarantee therefore applies to the discretized ODD, while this quantization determines the behavior between grid points.
This constitutes a deliberate deviation from the original table-based implementations, which interpolate the \(Q\)-values between grid vertices~\cite{Julian2019a, Julian2016}, and from the box-based verification~\cite{Damour2021}, which proves agreement over continuous regions; extending the point-wise guarantee to interval-based coverage of the continuous input space is left to future work.

The trained neural networks, lookup tables, and manifest files for both HCAS and VCAS are released alongside this paper~\cite{Christensen2026b}\footnote{\url{https://doi.org/10.5281/zenodo.20666415}} together with the code needed to reproduce the results\footnote{\url{https://github.com/DLR-KI/castrainer}}, constituting the first publicly available Safety Net implementation for these systems, and directly addressing a gap identified in prior work~\cite{Damour2021, Christensen2024}, where implementation details were not disclosed, precluding replicable results.

\section{Certification Impact}\label{sec:Certification}
This work is the first to systematically map Safety Nets to the learning assurance objectives of EASA's concept paper~\cite{EUASA2024}; \cref{tab:safety-net-objectives} enumerates the objectives addressed by the framework.
Here, three objectives are of particular significance.
Objective LM-04 requires quantifiable generalization bounds, and its anticipated MOC explicitly acknowledges that statistical learning theory bounds are typically too loose for large neural networks~\cite{EUASA2024}.
The Safety Net replaces this probabilistic argument with a deterministic one: by exhaustively evaluating the neural network against the MDP lookup tables across the entire discretized input space and storing all deviations, the generalization gap is identically zero by construction---a strictly stronger claim.
Objectives LM-10 and IMP-11, requiring requirements-based verification of the trained and inference models, respectively, follow directly: the exhaustive sweep constitutes verification against the formal specification for every valid input vector, satisfying the coverage requirements that DO-178C-aligned methods demand but that no finite test set can provide~\cite{EUASA2024, DO-178C}.
The remaining objectives listed in \cref{tab:safety-net-objectives}---including LM-09, LM-12 through LM-14, LM-16, IMP-09, IMP-10, and IMP-12---are addressed as structural by-products of the same sweep.

The mapping above refers to the published Issue~02~\cite{EUASA2024}.
Under the Proposed Issue~03~\cite{EUASA2026}, the three objectives of particular significance carry over without weakening the argument.
Objective LM-10 maps near-verbatim to Objective SU-LM-11, and Objective IMP-11 is covered by Objectives SU-IMP-06 and SU-IMP-07.
Objective LM-04 is split into an a priori and an a posteriori component, Objectives S-LM-06 and SU-LM-14, respectively.
This split is advantageous for Safety Nets:
while statistical learning theory bounds remain too loose to satisfy the a priori objective for large neural networks---the same limitation acknowledged for LM-04 in Issue~02---the deterministic, zero-error bound established by the exhaustive sweep satisfies the a posteriori objective SU-LM-14 by construction, as it directly verifies the generalization capability over the entire discretized input space rather than estimating it.
The remaining objectives carry over analogously: Objectives LM-09, LM-12, LM-13, and LM-14 map to Objectives SU-LM-10, SU-LM-12, SU-LM-13, and SU-LM-14; Objectives IMP-09 and IMP-10 are merged into Objective SU-IMP-07, and the completeness of Objectives LM-16 and IMP-12 move to Objectives SUR-DA-14 and SURK-DA-15.
A full re-mapping of \cref{tab:safety-net-objectives} to the Proposed Issue~03 convention is left to future work, pending the finalization of that document.

\section{Discussion}\label{sec:Discussion}
Compared to indications from prior literature~\cite{Dubey2019, Maas2013, Xu2015}, ReLU showed a clear and consistent superiority over LeakyReLU and GELU.
Given that the findings were derived for complex and smooth loss landscapes, it indicates that the discrete, bounded, piecewise-constant MDP policy underlying both HCAS and VCAS is missing these features.
ReLU's hard zeroing produces sparser activation patterns~\cite{Montufar2014} that align more closely with the sharp advisory boundaries of the target function.
In contrast, the non-zero negative slope of LeakyReLU introduces a gradient signal in regions where the policy is constant, and GELU's smooth nonlinearity offers no structural advantage here while incurring higher variance across runs.
The diminishing returns beyond 50 to 100 nodes per hidden layer are consistent with this interpretation: the representational complexity of the target function is bounded by the finite cardinality of the input space, and additional capacity is not exploited systematically beyond a moderate width, as evidenced by the substantially increased variance across training runs at higher widths.
The one-hot advantage observed in \cref{sec:PerformanceStudy} stands in apparent contrast to the prior Safety Net study, which found regression targets to outperform a classification formulation~\cite{Damour2021}.
Differences in architecture and encoding most plausibly explain the discrepancy: the prior study evaluated classification for only a single configuration---the wide, decreasing-width \emph{decreasing256} architecture designed for regression---whereas this paper sweeps narrow, uniform-width networks in which one-hot targets with cross-entropy loss align directly with the arg-max decision rule.
The two findings are therefore consistent with activation- and encoding-effects being strongly architecture- and task-dependent, reinforcing the need for the systematic sweep conducted here.

The hyperparameter study conducted in this work aggregates results across all subsystems of a given system, thereby obscuring subsystem-level variation and ignoring cross-coupling effects between hyperparameters at the individual subsystem level.
Moreover, a different split of HCAS and VCAS into alternative subsystems might be favorable for better compression.
A per-subsystem architecture search would likely yield a smaller combined system, with the greatest gains expected for VCAS, where subsystem-level variation in policy complexity is more pronounced, given the approximately two orders of magnitude difference in input space size relative to HCAS.
Similarly, the uniform layer width used throughout was not varied; decreasing-width architectures~\cite{Julian2016}, which could better match the effective dimensionality of the decision boundary as a function of depth, remain unexplored, as do architectures different from fully-connected neural networks.
The LUT still accounts for \qty{99.4}{\percent} of the combined VCAS system size---the primary reason the stated reduction relative to the monolithic \qty{4}{\gibi\byte} MDP lookup table reaches almost three orders of magnitude for HCAS but only around \num{1.25} orders of magnitude for VCAS.
However, a binary encoding, storing each entry as its \texttt{float32} input coordinates together with a single-byte advisory index, would require only \qty{17}{\byte} per entry for VCAS and \qty{13}{\byte} for HCAS---compared to the average footprint of approximately \qty{280}{\byte} per stored input vector for VCAS and \qty{210}{\byte} for HCAS in the k-d tree---reducing the LUT footprint by more than one order of magnitude without any loss of information.
Because the LUT dominates the combined VCAS system size, this serialization change alone would shrink the combined VCAS Safety Net from \qty{222.83}{\mebi\byte} to an estimated \qty{20}{\mebi\byte}, pushing the reduction relative to the monolithic \qty{4}{\gibi\byte} MDP lookup table beyond two orders of magnitude for both systems.
Further headroom remains through packed grid indices instead of \texttt{float32} coordinates and through established compression techniques such as pruning, quantization, and Huffman coding~\cite{Han2016, Gholami2022}, the latter also suggested for the hybrid architecture in~\cite{Damour2021}; these optimizations, however, are left to future work, as the present implementation prioritizes the auditable manifest format over minimal footprint.
Finally, the open-source HCAS and VCAS implementations are proof-of-concept approximations of ACAS~Xa and ACAS~Xu that do not cover the full operational specifications of ED-256/DO-385 and ED-275/DO-386~\cite{ED-256, ED-275, DO-385, DO-386}; the reported system sizes should therefore not be interpreted as representative of a production deployment.
In fact, the gap is substantial along several axes, as summarized in \cref{tab:spec-coverage}.
HCAS fixes both ownship and intruder speed to a single value of \qty{200}{\foot\per\second}, collapsing two state dimensions that the standards define over the full \qtyrange{0}{600}{\knot} envelope, and truncates range at \qty{56000}{\foot} against the \qty{14}{\nauticalmile} (\(\approx\)~\qty{85000}{\foot}) head-on surveillance requirement~\cite{DO-386}.
VCAS clips vertical rates at \qty{\pm100}{\foot\per\second} (\qty{\pm6000}{\foot\per\minute}) against the \qty{\pm10000}{\foot\per\minute} design maximum~\cite{DO-385}, and both systems collapse the previous-advisory state to a single index rather than the full sense/strength/crossing/coordination structure of the standardized resolution advisory encodings.
Discretizing the full operational envelopes at strides comparable to the open-source tables yields on the order of \num{1.2e9} states for ACAS~Xu and \num{1.3e8} for ACAS~Xa per subsystem partition---roughly four and one and a half orders of magnitude larger, respectively, than the \num{53792} and \num{4053465} points of a single HCAS or VCAS subsystem.
Because the LUT footprint of a Safety Net scales with the number of misrepresented input vectors, and the neural network's footprint scales with the representational complexity of a far richer policy, a production-scale Safety Net would be considerably larger than the figures in \cref{tab:SafetyNetSizes}; the present results should be read as a methodological demonstration on a faithful but reduced surrogate, not as a size estimate for certified ACAS~X.

Furthermore, the results bear directly on the certification argument.
Because the Safety Net guarantees \qty{100}{\percent} agreement with the lookup table ground truth across the entire discretized input space by construction, the architectural choices studied here---activation function, depth, width, and encoding---do not affect whether the EASA verification objectives are met, but only the cost at which they are met.
A higher neural network retrieval rate shrinks the lookup table, and therefore the combined system size, but even the poorest-performing configuration remains certifiable, as the lookup table absorbs every residual error.
In this sense, the hyperparameter study optimizes the deployability of a system that is correct by design, rather than trading accuracy against safety: the learning assurance and implementation objectives enumerated in \cref{tab:safety-net-objectives}---in particular, the requirements-based verification objectives LM-10 and IMP-11 and the generalization objective LM-04---are satisfied identically across all evaluated architectures.

\section{Conclusion}\label{sec:Conclusion}
This work presents the first open-source and reproducible set of Safety Nets together with a systematic analysis of Safety Nets as a Safety-by-Design solution for certifiable neural networks in aviation, systematically varying activation function, depth, width, and output encoding across all 40 HCAS and 9 VCAS subsystems.
Contrary to the general preference in recent literature~\cite{Dubey2019, Maas2013, Xu2015}, ReLU matches or outperforms LeakyReLU and GELU across both systems---attributable to the piecewise-constant structure of the underlying MDP policy, for which ReLU's sparse activation patterns are particularly well-suited.
The optimal architecture lies in the range of 3 to 5 hidden layers with approximately 50 to 100 nodes per layer; beyond this, variance increases, and the neural network's memory footprint grows without a corresponding reduction in LUT size, increasing the combined system size.
Furthermore, one-hot encoding consistently outperforms target encoding by at least one order of magnitude on the median LUT percentage, particularly for smaller architectures.
The resulting Safety Nets achieve average neural network coverages of \qty{99.65}{\percent} and \qty{97.77}{\percent} for HCAS and VCAS, with combined system sizes of \qty{4.76}{\mebi\byte} and \qty{222.83}{\mebi\byte}, respectively---reductions of almost three and slightly more than one order of magnitude relative to the monolithic \qty{4}{\gibi\byte} MDP lookup table.
The Safety Net framework satisfies a set of EASA learning assurance and implementation objectives by construction, replacing probabilistic generalization arguments with a deterministic, zero-error bound over the complete discretized input space and thereby fulfilling requirements-based verification objectives LM-10 and IMP-11 without a separate post-training verification campaign, as detailed in \cref{tab:safety-net-objectives}.
Crucially, these objectives are satisfied by construction regardless of the chosen architecture, so the hyperparameter optimization presented here serves to minimize deployment cost on avionics hardware rather than to trade accuracy against certifiability.
Finally, the trained neural networks, lookup tables, manifest files, and training scripts for both HCAS and VCAS are released alongside this paper, constituting the first publicly available and reproducible Safety Net implementation.

\section{Contact Author Email Address}
\underline{\href{mailto:johann.christensen@dlr.de}{johann.christensen@dlr.de}}

\section{Copyright Statement}
\begin{small}
    The authors confirm that they, and/or their company or organization, hold copyright on all of the original material included in this paper. The authors also confirm that they have obtained permission, from the copyright holder of any third party material included in this paper, to publish it as part of their paper. The authors confirm that they give permission, or have obtained permission from the copyright holder of this paper, for the publication and distribution of this paper as part of the ICAS proceedings or as individual off-prints from the proceedings.
\end{small}

\FloatBarrier{}

\biblio{literature-bibtex}\label{bibliography}

\clearpage

\appendix

\section{Safety Net Manifests}
\begin{figure}[htb]
    \centering
\begin{lstlisting}[language=json]
{
  "version": "1.0.0",
  "description": "SafetyNet for vertical collision avoidance in aircraft. Provides 9 advisories: COC, DNC, DND, DES1500, CL1500, SDES1500, SCL1500, SDES2500, SCL2500.",
  "function": "Vertical Collision Avoidance System",
  "datatype": "float32",
  "inputs": [
    {
      "id": "h",
      "name": "Relative altitude",
      "description": "Relative altitude between ownship and intruder",
      "unit": "ft",
      "ranges": [
        { "minimum": -8000.0, "maximum": -4000.0, "stride": 1000.0 },
        ...
      ]
    },
    ...
  ],
  "numberOutputs": 9,
  "networks": [
    {
      "file": "vcas_01.pt",
      "networkFormat": "torch",
      "if": {
        "s_adv": { "minimum": 1.0, "maximum": 1.0 }
      }
    },
    ...
  ],
  "luts": [
    {
      "file": "vcas_01_lut.json",
      "lutFormat": "snet",
      "if": {
        "s_adv": { "minimum": 1.0, "maximum": 1.0 }
      }
    },
    ...
  ]
}
\end{lstlisting}%
    \caption{General structure of the Safety Net manifest for VCAS.}
    \label{fig:SNM_VCAS}
\end{figure}
\begin{figure}[htb]
    \centering
\begin{lstlisting}[language=json]
{
  "version": "1.0.0",
  "description": "SafetyNet for horizontal collision avoidance in aircraft. Provides 5 advisories: COC, WL, WR, SL, SR.",
  "function": "Horizontal Collision Avoidance System",
  "datatype": "float32",
  "inputs": [
    {
      "id": "rho",
      "name": "Range",
      "description": "Range to intruder",
      "unit": "ft",
      "ranges": [
        { "minimum": 0.0, "maximum": 100.0, "stride": 25.0 },
        ...
      ]
    },
    ...
  ],
  "numberOutputs": 5,
  "networks": [
    {
      "file": "hcas_pra0_tau00.pt",
      "networkFormat": "torch",
      "if": {
        "s_adv": { "minimum": 0.0, "maximum": 0.0 },
        "tau": { "minimum": 0.0, "maximum": 0.0 }
      }
    },
    ...
  ],
  "luts": [
    {
      "file": "hcas_pra0_tau00_lut.json",
      "lutFormat": "snet",
      "if": {
        "s_adv": { "minimum": 0.0, "maximum": 0.0 },
        "tau": { "minimum": 0.0, "maximum": 0.0 }
      }
    },
    ...
  ]
}
\end{lstlisting}%
    \caption{General structure of the Safety Net manifest for HCAS.}
    \label{fig:SNM_HCAS}
\end{figure}

\FloatBarrier{}

\section{Full Training Configuration for HCAS and VCAS}\label{sec:HCASVCASConfig}
\begin{table}[htb]
    \caption{%
        Discretized input space of the VCAS SafetyNet.%
    }\label{tab:VCAS_grid}
    \centering
    \begin{tabular}{lllll}
        \toprule
        Variable                 & Description            & Range                                                           & Stride                         & Comment \\
        \midrule
        \(h\)                    & Relative altitude      & [\qty{-8000}{\foot}, \qty{-4000}{\foot}]                    & \qty{1000}{\foot}          &\\
                                 &                        & [\qty{-3000}{\foot}, \qty{-1250}{\foot}]                    & \qty{250}{\foot}         &\\
                                 &                        & [\qty{-1000}{\foot}, \qty{-800}{\foot}]                     & \qty{100}{\foot}           &\\
                                 &                        & [\qty{-700}{\foot}, \qty{-150}{\foot}]                      & \qty{50}{\foot}            &\\
                                 &                        & [\qty{-100}{\foot}, \qty{100}{\foot}]                       & \qty{25}{\foot}           &\\
                                 &                        & [\qty{150}{\foot}, \qty{700}{\foot}]                        & \qty{50}{\foot}            &\\
                                 &                        & [\qty{800}{\foot}, \qty{1000}{\foot}]                       & \qty{100}{\foot}          &\\
                                 &                        & [\qty{1250}{\foot}, \qty{3000}{\foot}]                      & \qty{250}{\foot}           &\\
                                 &                        & [\qty{4000}{\foot}, \qty{8000}{\foot}]                      & \qty{1000}{\foot}          &\\
        \(\dot{h}_\mathrm{own}\) & Ownship vertical rate  & [\qty{-100}{\foot\per\second}, \qty{-60}{\foot\per\second}] & \qty{10}{\foot\per\second} &\\
                                 &                        & [\qty{-50}{\foot\per\second}, \qty{-35}{\foot\per\second}]  & \qty{5}{\foot\per\second}  &\\
                                 &                        & [\qty{-30}{\foot\per\second}, \qty{30}{\foot\per\second}]   & \qty{3}{\foot\per\second}  &\\
                                 &                        & [\qty{35}{\foot\per\second}, \qty{50}{\foot\per\second}]    & \qty{5}{\foot\per\second}  &\\
                                 &                        & [\qty{60}{\foot\per\second}, \qty{100}{\foot\per\second}]   & \qty{10}{\foot\per\second} &\\
        \(\dot{h}_\mathrm{int}\) & Intruder vertical rate & [\qty{-100}{\foot\per\second}, \qty{-60}{\foot\per\second}] & \qty{10}{\foot\per\second} &\\
                                 &                        & [\qty{-50}{\foot\per\second}, \qty{-35}{\foot\per\second}]  & \qty{5}{\foot\per\second}  &\\
                                 &                        & [\qty{-30}{\foot\per\second}, \qty{30}{\foot\per\second}]   & \qty{3}{\foot\per\second}  &\\
                                 &                        & [\qty{35}{\foot\per\second}, \qty{50}{\foot\per\second}]    & \qty{5}{\foot\per\second}  &\\
                                 &                        & [\qty{60}{\foot\per\second}, \qty{100}{\foot\per\second}]   & \qty{10}{\foot\per\second} &\\
        \(\tau{}\)               & Time to CPA            & [\qty{0}{\second}, \qty{40}{\second}]                       & \qty{1}{\second}           &\\
        \(s_\mathrm{adv}\)       & Previous advisory      & [0, 8]                                                      & \num{1}                    &\\
        \bottomrule
    \end{tabular}
\end{table}
\begin{table}[htb]
    \caption{%
        Discretized input space of the HCAS SafetyNet.%
    }\label{tab:HCAS_grid}
    \centering
    \begin{tabular}{lllll}
        \toprule
        Variable           & Description       & Range                                                          & Stride                                       &Comment \\
        \midrule
        \(\rho{}\)         & Range to intruder & [\qty{0}{\foot}, \qty{100}{\foot}]                         & \qty{25}{\foot}                          & \\
                           &                   & [\qty{150}{\foot}, \qty{200}{\foot}]                       & \qty{50}{\foot}                          & \\
                           &                   & [\qty{300}{\foot}, \qty{500}{\foot}]                       & \qty{100}{\foot}                         & \\
                           &                   & [\qty{510}{\foot}, \qty{510}{\foot}]                       & \qty{1}{\foot}     & single point\\
                           &                   & [\qty{750}{\foot}, \qty{1000}{\foot}]                      & \qty{250}{\foot}                         & \\
                           &                   & [\qty{1500}{\foot}, \qty{2000}{\foot}]                     & \qty{500}{\foot}                         & \\
                           &                   & [\qty{3000}{\foot}, \qty{5000}{\foot}]                     & \qty{1000}{\foot}                        & \\
                           &                   & [\qty{7000}{\foot}, \qty{13000}{\foot}]                    & \qty{2000}{\foot}                        & \\
                           &                   & [\qty{15000}{\foot}, \qty{21000}{\foot}]                   & \qty{2000}{\foot}                        & \\
                           &                   & [\qty{25000}{\foot}, \qty{40000}{\foot}]                   & \qty{5000}{\foot}                        & \\
                           &                   & [\qty{48000}{\foot}, \qty{56000}{\foot}]                   & \qty{8000}{\foot}                        & \\
        \(\theta{}\)       & Bearing angle     & [\(-\pi\), \(\pi\)]                                                & \(\sfrac{\pi}{20}\)                          & \\
        \(\psi{}\)         & Relative heading  & [\(-\pi\), \(\pi\)]                                                & \(\sfrac{\pi}{20}\)                          & \\
        \(v_\mathrm{own}\) & Ownship speed     & [\qty{200}{\foot\per\second}, \qty{200}{\foot\per\second}] & \qty{1}{\foot\per\second} & fixed \\
        \(v_\mathrm{int}\) & Intruder speed    & [\qty{200}{\foot\per\second}, \qty{200}{\foot\per\second}] & \qty{1}{\foot\per\second}  & fixed\\
        \(\tau{}\)         & Time to CPA       & [\qty{0}{\second}, \qty{15}{\second}]                      & \qty{5}{\second}                         & \\
                           &                   & [\qty{20}{\second}, \qty{40}{\second}]                     & \qty{10}{\second}                        & \\
                           &                   & [\qty{60}{\second}, \qty{60}{\second}]                     & \qty{1}{\second}    & single point\\
       \( s_\mathrm{adv}\) & Previous advisory & [0, 4]                                                     & \num{1}                                  & \\
        \bottomrule
    \end{tabular}
\end{table}

\FloatBarrier{}

\section{Specification Coverage of HCAS and VCAS}\label{sec:SpecCoverage}
\begin{longtblr}[
    caption = {Parameter and resolution advisory (RA) coverage of the open-source HCAS/VCAS use case~\cite{Julian2019} relative to the full ACAS~Xu/Xa specifications~\cite{ED-256, ED-275, DO-385, DO-386}. Speed envelopes and the \qty{14}{\nauticalmile} surveillance range are taken from DO-386 \S{}2.2.1.3 and \S{}2.2.2.1.1; the \qty{\pm10000}{\foot\per\minute} vertical-rate design maximum from DO-385 \S{}2.2.4.6; and the RA strength and horizontal turn encodings from the DO-385 Strength-Bits table and the DO-386 horizontal sense (Turn Left/Right and target-track-angle) fields, respectively.},
    label = {tab:spec-coverage},
    ]{
    colspec = {lX[2,l]X[2,l]X[1,l]},
    rowhead = 1,
    rowfoot = 0,
    }
    \toprule
    \textbf{Variable}                                                                 & \textbf{HCAS/VCAS}                                                       & \textbf{DO-385/DO-386}                                                                                                                                              & \textbf{Covered}               \\
    \midrule
    \SetCell[c=4]{l}\emph{ACAS~Xu/HCAS}                                               &                                                                          &                                                                                                                                                                     &                                \\
    \midrule
    Range \(\rho{}\)                                                                  & [\qty{0}{\foot}, \qty{56000}{\foot}]                                 & up to \qty{14}{\nauticalmile} (\(\approx\)~\qty{85000}{\foot}) head-on; \(\geq\)~\qty{1000}{\foot} min.\ track                                                        & partial                        \\
    Bearing \(\theta{}\)                                                              & [\ang{-180}, \ang{180}]                                              & [\ang{-180}, \ang{180}]                                                                                                                                         & full                           \\
    Rel.\ heading \(\psi{}\)                                                          & [\ang{-180}, \ang{180}]                                              & [\ang{-180}, \ang{180}]                                                                                                                                         & full                           \\
    Ownship speed \(v_\mathrm{own}\)                                                  & \qty{200}{\foot\per\second} (fixed)                                      & \qtyrange{0}{600}{\knot} (\(\approx\)~\qtyrange{0}{1013}{\foot\per\second})                                                                                          & none                           \\
    Intruder speed \(v_\mathrm{int}\)                                                 & \qty{200}{\foot\per\second} (fixed)                                      & \qtyrange{0}{600}{\knot}                                                                                                                                            & none                           \\
    Closing speed                                                                     & \(\approx\)~\qty{400}{\foot\per\second} (implied)                         & up to \qty{1200}{\knot} head-on (\(\approx\)~\qty{2025}{\foot\per\second})                                                                                           & partial                        \\
    Time to CPA \(\tau{}\)                                                            & \{0,5,10,15,20,30,40,60\}\,\unit{\second}                           & continuous, comparable horizon                                                                                                                                      & sampled                        \\
    Prev.\ advisory \(s_\mathrm{adv}\)                                                & \{0,\dots,4\} (index)                                                  & turn sense + 6-bit target track angle + reversals + coordination                                                                                                    & partial                        \\
    Vertical dimension                                                                & none                                                                     & blended vertical + horizontal RAs                                                                                                                                 & none                           \\
    RA set                                                                            & COC, WL, WR, SL, SR                                                 & Turn Left/Right toward target track angle, reversals, RWC bands                                                                                                     & proxy                          \\
    \midrule
    \SetCell[c=4]{l}\emph{ACAS~Xa/VCAS}                                               &                                                                          &                                                                                                                                                                     &                                \\
    \midrule
    Rel.\ altitude \(h\)                                                              & [\qty{-8000}{\foot}, \qty{8000}{\foot}]                              & \qty{\pm10000}{\foot} surveillance; \qty{\pm3000}{\foot} core RA region                                                                                             & partial                        \\
    Ownship vert.\ rate \(\dot{h}_\mathrm{own}\)                                      & \qty{\pm100}{\foot\per\second} (\qty{\pm6000}{\foot\per\minute})         & \qty{\pm10000}{\foot\per\minute} (\qty{\pm167}{\foot\per\second}) design max                                                                                        & partial                        \\
    Intruder vert.\ rate \(\dot{h}_\mathrm{int}\)                                     & \qty{\pm100}{\foot\per\second}                                           & \qty{\pm10000}{\foot\per\minute}                                                                                                                                   & partial                        \\
    Time to CPA \(\tau{}\)                                                            & [\qty{0}{\second}, \qty{40}{\second}]                                & comparable horizon (to \(\approx\)~\qty{60}{\second})                                                                                                                & partial                        \\
    Prev.\ advisory \(s_\mathrm{adv}\)                                                & \{0,\dots,8\} (index)                                                  & sense + strength + crossing + coordination state                                                                                                                    & partial                        \\
    RA set                                                                            & COC, DNC, DND, DES1500, CL1500, SDES1500, SCL1500, SDES2500, SCL2500 & full Strength-Bits set: Monitor~VS, Level-Off, \num{\pm1500}, Increase-Rate, Maintain-Rate (\textgreater\qty{1500}{\foot\per\minute}), Reversals, MTLO (approx.\ 14 codes) & partial                        \\
    \midrule
    \SetCell[c=3]{l}\emph{Approx.\ discretized state count (per subsystem partition)} &                                                                          &                                                                                                                                                                     & \textbf{Difference}                               \\
    \midrule
    HCAS/ACAS~Xu                                                                      & \num{53792}                                                               & approx.\ \num{1.2e9} (full-envelope grid)                                                                                                                           & approx.\ 4 orders of magnitude \\
    VCAS/ACAS~Xa                                                                      & \num{4053465}                                                             & approx.\ \num{1.3e8} (full-envelope grid)                                                                                                                           & approx.\ 1.5 orders of magnitude \\
    \bottomrule
\end{longtblr}

\FloatBarrier{}

\section{EASA Objectives Addressed by Safety Nets}\label{sec:ObjectiveSafetyNets}
\begin{longtblr}[
        caption = {Objectives from the EASA concept paper fulfilled or partially addressed by the Safety Net framework~\cite{Christensen2024}.},
        label = {tab:safety-net-objectives},
    ]{
        colspec = {lXX},
        rowhead = 1,
        rowfoot = 0,
    }

    \toprule
    \textbf{Objective} &
    \textbf{Excerpt}   &
    \textbf{Rationale}   \\
    \midrule

    LM-04
                       &
    {%
            \enquote{The applicant should provide quantifiable generalization bounds.}
    \\
            \emph{Comment: the anticipated MOC notes that bounds from statistical learning theory are often too loose for large neural networks without unreasonable amounts of training data.}
        }
                       &
    By exhaustively evaluating the neural network against the formal ground truth (e.g., the Markov decision process tables) across the entire discretized input space, the Safety Net guarantees a generalization gap of exactly zero for all valid, discretized input vectors.
    This is a deterministic, non-probabilistic bound---stronger than any statistical learning theory-based bound---and satisfies the objective by construction whenever the input space can be meaningfully discretized.
    \\

    \midrule

    LM-09
                       &
    \enquote{The applicant should perform an evaluation of the performance of the trained model based on the test data set and document the result of the model verification.}
                       &
    The Safety Net construction process evaluates the trained model on the complete discretized input space rather than on a sampled test set.
    Every deviation from the ground truth is detected and recorded.
    The resulting performance figure is not estimated but exact, providing stronger evidence than a finite test set could supply, and all deviations are explicitly documented in the sparse lookup table.
    \\

    \midrule

    LM-10
                       &
    \enquote{The applicant should perform requirements-based verification of the trained model behavior.}
                       &
    Each entry in the Safety Net represents a point at which the neural network's output was verified against the formal specification (ground truth).
    The full sweep constitutes exhaustive requirements-based verification: every input vector is checked against the decision rule, coverage of all requirements by test cases is guaranteed, and all discrepancies are explicitly stored and corrected.
    \\

    \midrule

    LM-12
                       &
    {%
            \enquote{The applicant should perform and document the verification of the stability of the trained model, covering the whole AI/ML constituent ODD.}
    \\
            \emph{Comment: the anticipated MOC requires coverage of nominal cases as well as singular points, edge cases, and corner cases.}
        }
                       &
    The exhaustive sweep of the Safety Net construction inherently covers all of these, since every point in the discretized ODD is evaluated.
    No separate, targeted stability test suite is needed; stability evidence is a by-product of building the Safety Net.
    \\

    \midrule

    LM-13
                       &
    {%
            \enquote{The applicant should perform and document the verification of the robustness of the trained model in adverse conditions.}
    \\
            \emph{Comment: the anticipated MOC lists singular points, edge and corner cases within the ODD, and out-of-distribution test cases as required evidence.}
        }
                       &
    Exhaustive coverage of the discretized input space subsumes all in-ODD corner and edge cases.
    Because the training data comprise the complete discretized ODD, no distributional shift between training and operation can occur within the ODD; out-of-distribution inputs can only arise from states outside the ODD bounds, which are deterministically detectable through range checks against the manifest's input specification and must be mitigated at the (sub)system level.
    While formal methods are cited as a promising MOC, the Safety Net approach is functionally equivalent to exhaustive formal model checking over the discretized domain.
    \\

    \midrule

    LM-14
                       &
    \enquote{The applicant should verify the anticipated generalization bounds using the test data set.}
                       &
    Since the Safety Net has already verified the model's output for every discretized input point (satisfying LM-04 with a zero-error bound), LM-14 is satisfied automatically: the entire discretized input space serves as the verification dataset, and the absence of any uncorrected deviation in the sparse lookup table confirms the zero generalization gap.
    \\

    \midrule

    LM-16
                       &
    {%
            \enquote{The applicant should confirm that the trained model verification activities are complete.}
    \\
            \emph{Comment: the anticipated MOC requires coverage of all pairs of ODD parameters and coverage of the whole AI/ML constituent ODD for stability.}
        }
                       &
    The discretized input count and the parallel sweep time can be calculated exactly, providing a traceable, quantitative completeness argument.
    The systematic manifest format and business logic checks further document that all neural networks and lookup tables cover the required input space without gaps or overlaps (objectives G-004, G-007, G-008 of the manifest schema~\cite{Christensen2024}).
    \\

    \midrule

    IMP-09
                       &
    {%
            \enquote{The applicant should perform and document the verification of the stability of the inference model.}
    \\
            \emph{Comment: the anticipated MOC requires coverage of nominal cases, singular points, edge cases, and corner cases throughout the ODD.}
        }
                       &
    Equivalent to LM-12 but applied to the inference model: full discretized ODD coverage during the Safety Net sweep implicitly covers all stability cases.
    \\

    \midrule

    IMP-10
                       &
    \enquote{The applicant should perform and document the verification of the robustness of the inference model in adverse conditions.}
                       &
    Equivalent to LM-13 at the inference model level.
    The exhaustive sweep covers all in-ODD edge and corner cases.
    \\

    \midrule

    IMP-11
                       &
    {%
            \enquote{The applicant should perform requirements-based verification of the inference model behavior when integrated into the AI/ML constituent.}
    \\
            \emph{Comment: the anticipated MOC requires that verification cases cover all requirements allocated to the AI/ML constituent.}
        }
                       &
    The complete discretized input sweep against the formal ground truth constitutes exhaustive requirements-based verification of the integrated inference model.
    The manifest's condition system (wildcards, ranges, strides) ensures that every allowed input vector is covered by exactly one network or lookup table, providing a formal coverage argument at the constituent level.
    \\

    \midrule

    IMP-12
                       &
    \enquote{The applicant should confirm that the AI/ML constituent verification activities are complete.}
                       &
    Analogous to LM-16 at the implementation level.
    The quantitative completeness argument (total input count, sweep time, zero residual errors after Safety Net application) provides traceable evidence.
    The JSON manifest and its associated business logic checks (cf.\ Table~II in~\cite{Christensen2024}) create an auditable record that all constituent requirements have been verified and all discrepancies corrected.
    \\

    \bottomrule

\end{longtblr}

\FloatBarrier{}

\end{document}